%% file: main.tex
\documentclass[runningheads]{llncs}
\usepackage{eccv}

\usepackage{eccvabbrv}

\usepackage{graphicx}
\usepackage{booktabs}
\usepackage{multirow}
\usepackage[numbers,sort&compress,sectionbib]{natbib}

\usepackage[table]{xcolor}
\definecolor{baselinegray}{gray}{0.93}
\definecolor{navyblue}{RGB}{0,0,128}
\definecolor{vegablue}{RGB}{65, 105, 225}

\providecommand{\model}[1]{\textbf{#1}} 
\usepackage[accsupp]{axessibility}  
\usepackage{makecell}

\usepackage{bm}
\usepackage{amsmath}
\usepackage{amssymb}
\usepackage{wrapfig}
\usepackage{caption}
\usepackage{algorithm}
\usepackage{algpseudocode}
\newcommand{\myparagraph}[1]{\vspace{3pt}\noindent\textbf{#1}}

\newcommand{\squishlist}{
   \begin{list}{$\bullet$}
    { \setlength{\itemsep}{0pt}
      \setlength{\parsep}{0pt}
      \setlength{\topsep}{0pt}
      \setlength{\partopsep}{0pt}
      \setlength{\leftmargin}{1.0em} \setlength{\labelwidth}{1em}
      \setlength{\labelsep}{0.5em} } }
      
\newcommand{\squishend}{
    \end{list}}

\usepackage{hyperref}

\usepackage{orcidlink}
\usepackage{marvosym}
\begin{document}
\renewcommand{\bibname}{References}
\title{Generation Models Know Space: Unleashing Implicit 3D Priors for Scene Understanding}

\titlerunning{Generation Models Know Space: VEGA-3D}

\author{Xianjin Wu\inst{1} \and
Dingkang Liang\inst{1}\textsuperscript{\dag} \and
Tianrui Feng\inst{1} \and
Kui Xia\inst{2} \and
Yumeng Zhang\inst{2} \and
Xiaofan Li\inst{2} \and
Xiao Tan\inst{2} \and
Xiang Bai\inst{1}\textsuperscript{\Letter}
}

\authorrunning{X.Wu et al.}

\institute{Huazhong University of Science and Technology, China \and
Baidu Inc., China \\
\textsuperscript{\dag}\,Project Lead \quad \textsuperscript{\Letter}\,Corresponding author \\
\email{\{wuxianjin,dkliang,xbai\}@hust.edu.cn}}

\maketitle

\vspace{-20pt}
\begin{figure*}[h]
\centering
\includegraphics[width=0.99\textwidth]{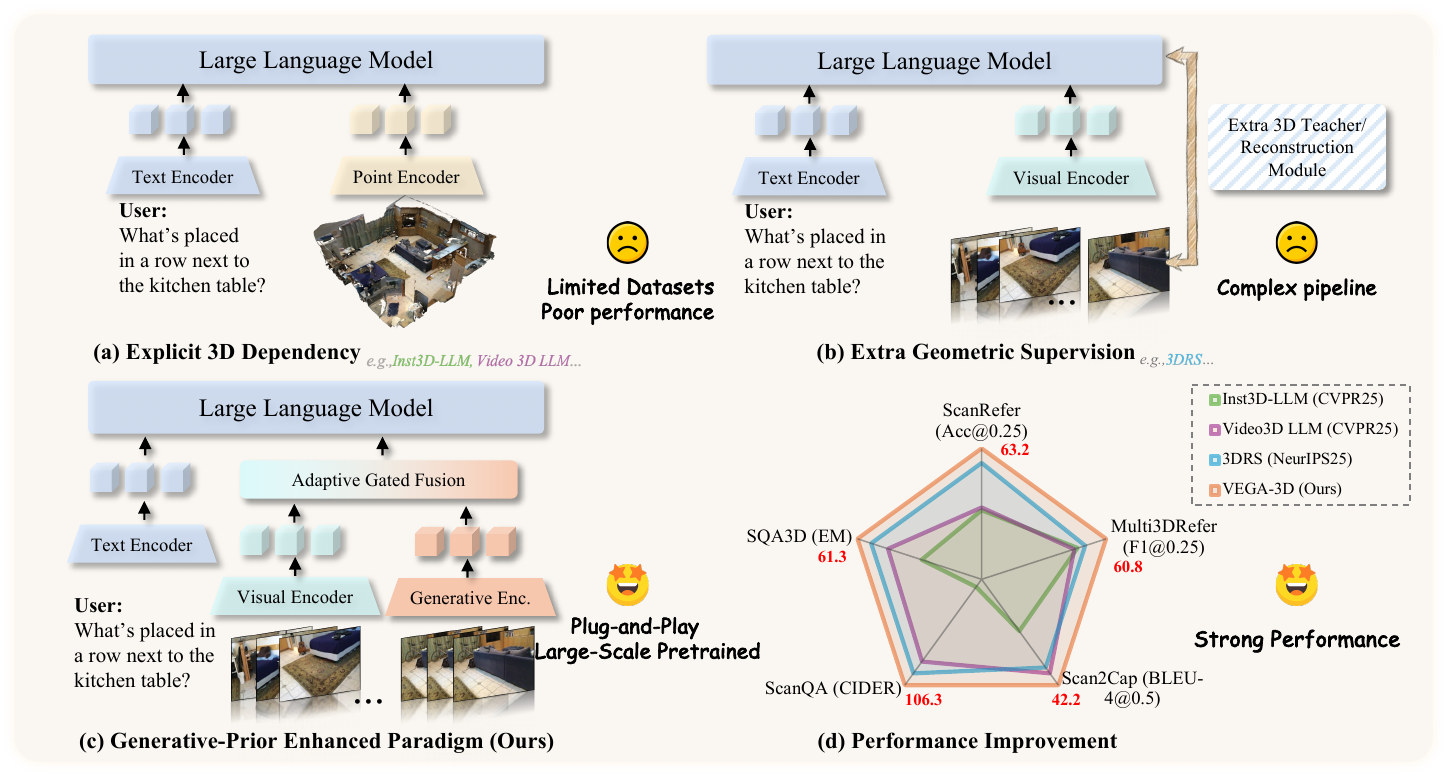}
\vspace{-10pt}
\caption{\textbf{Comparison of existing paradigms.} Unlike methods relying on (a) explicit 3D inputs or (b) complex geometric supervision, (c) our \model{VEGA-3D} extracts implicit priors from video generation models. By repurposing them as Latent World Simulators, we achieve (d) superior performance without external 3D dependencies.}
\vspace{-10pt}
\label{fig: intro}
\end{figure*}

\vspace{-20pt}

\begin{abstract}
While Multimodal Large Language Models demonstrate impressive semantic capabilities, they often suffer from \textbf{spatial blindness}, struggling with fine-grained geometric reasoning and physical dynamics. Existing solutions typically rely on explicit 3D modalities or complex geometric scaffolding, which are limited by data scarcity and generalization challenges. In this work, we propose a paradigm shift by leveraging the implicit spatial prior within large-scale video generation models. We posit that to synthesize temporally coherent videos, these models inherently learn robust 3D structural priors and physical laws. We introduce \model{VEGA-3D} (\textbf{V}ideo \textbf{E}xtracted \textbf{G}enerative \textbf{A}wareness), a plug-and-play framework that repurposes a pre-trained video diffusion model as a Latent World Simulator. By extracting spatiotemporal features from intermediate noise levels and integrating them with semantic representations via a token-level adaptive gated fusion mechanism, we enrich MLLMs with dense geometric cues without explicit 3D supervision. Extensive experiments across 3D scene understanding, spatial reasoning, and embodied manipulation benchmarks demonstrate that our method outperforms state-of-the-art baselines, validating that generative priors provide a scalable foundation for physical-world understanding. Code is publicly available at \url{https://github.com/H-EmbodVis/VEGA-3D}.

\end{abstract}

\section{Introduction}

Recent advancements in video generation models \cite{wan2025wan, genie3, huang2025vid2world, li2025vmem, jiang2025vace} have reshaped our expectations of visual systems, moving beyond high-fidelity generation to acting as interactive world models \cite{kang2024far, valevski2024diffusion, xiao2025worldmem, xu2026nextforcing}. To generate a plausible video, the model inherently aligns appearance with 3D geometry: occlusion requires persistent object identity, camera motion reveals depth-dependent apparent motion, and interactions must follow consistent dynamics. These constraints encourage latent representations that encode geometry-consistent structure and motion, yielding a strong learned 3D prior without explicit 3D supervision \cite{ren2025gen3c, kim2025videofrom3d}. This raises a compelling research question: if video generators already possess an implicit understanding of space and physics, can these implicit physical priors be repurposed to improve downstream 3D visual understanding?

This perspective is particularly critical for domains that require granular 3D awareness \cite{gao2024physically, cheang2024gr, zheng2024towards, xu2024unified, liu2025embodied, Zhang_2026_CVPR, fang2026towards, fu2025minddrive}, such as scene understanding. To equip embodied agents with such capabilities, prevailing research has predominantly followed two explicit paradigms, as illustrated in Fig.~\ref{fig: intro}. The first stream~\cite{pointllm, video3dllm} directly utilizes explicit 3D modalities (\eg, point clouds or depth) to provide definitive geometric grounding.  The second stream \cite{huang2025, wang2025ross3d, chen2025think} focuses on geometric scaffolding, which lifts 2D features into 3D space via extra reconstruction or distillation. Alongside these methods, an underexplored yet increasingly promising paradigm (Fig.~\ref{fig: intro}(c)) lies in modern video generation models trained on large-scale video datasets, whose training objective implicitly rewards representations consistent with 3D geometry and physical dynamics.

\begin{figure*}[t]
\centering
\includegraphics[width=0.99\textwidth]{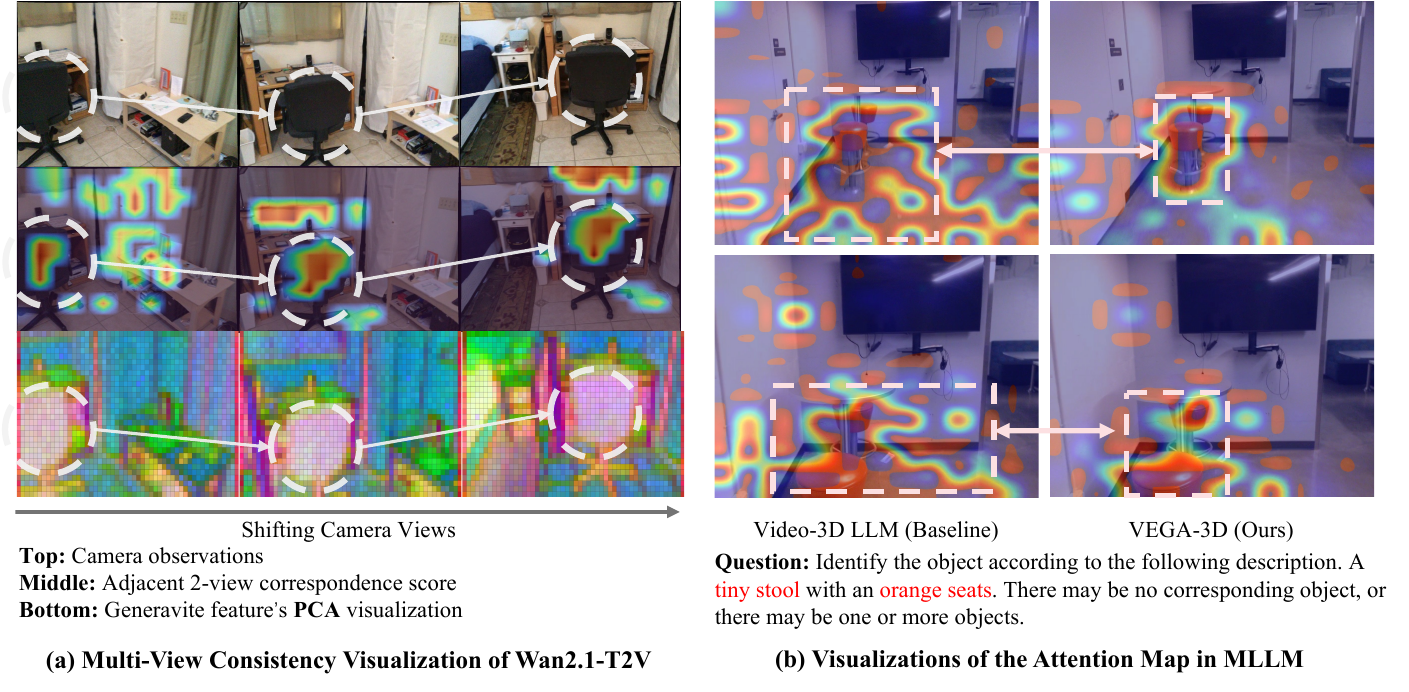}
\vspace{-5pt}
\caption{\textbf{Visualization of implicit 3D priors.} (a) The generation model demonstrates strong multi-view geometric consistency, evidenced by high correspondence scores and stable PCA feature representations across shifting camera views. (b) By leveraging these priors, our \model{VEGA-3D} overcomes the spatial ambiguity observed in the baseline, yielding precisely-located attention maps of the target object in the instruction.}
\vspace{-15pt}
\label{fig: consistency&attn}
\end{figure*}

In this work, we explore a new paradigm: leveraging representations learned by video generation models as priors for geometric understanding. As illustrated in the Fig.~\ref{fig: consistency&attn} (a), video diffusion models demonstrate remarkable multi-view consistency. The model captures the structural integrity of objects across different frames, implying a robust internal representation of 3D geometry. While generative models lack the semantic alignment of contrastive pre-training \cite{zhai2023sigmoid, siglip2}, their geometric priors offer unique spatial guidance. As further evidenced in Fig.~\ref{fig: consistency&attn}(b), incorporating these priors sharpens the original scattered attention of the baseline, effectively serving as spatial anchors that enable precise localization for fine-grained 3D reasoning.

Motivated by these observations, we propose \model{VEGA-3D}, a plug-and-play framework that incorporates the strengths of semantic and generative representations. Specifically, we introduce a video generation model (e.g., Wan2.1~\cite{ wan2025wan}, Vmem~\cite{li2025vmem}) as a \textit{Latent World Simulator} to enrich the visual stream with spatiotemporal world-knowledge priors, complementary to the semantic encoder. To solve the distribution shift between generative and semantic space, we design a token-level adaptive gated fusion module that integrates the two features. This fusion enables the model to actively exploit the generative backbone's 3D awareness to strengthen geometry-sensitive reasoning, while preserving discriminative semantic cues.

\begin{figure*}[t]
\centering
\includegraphics[width=0.99\textwidth]{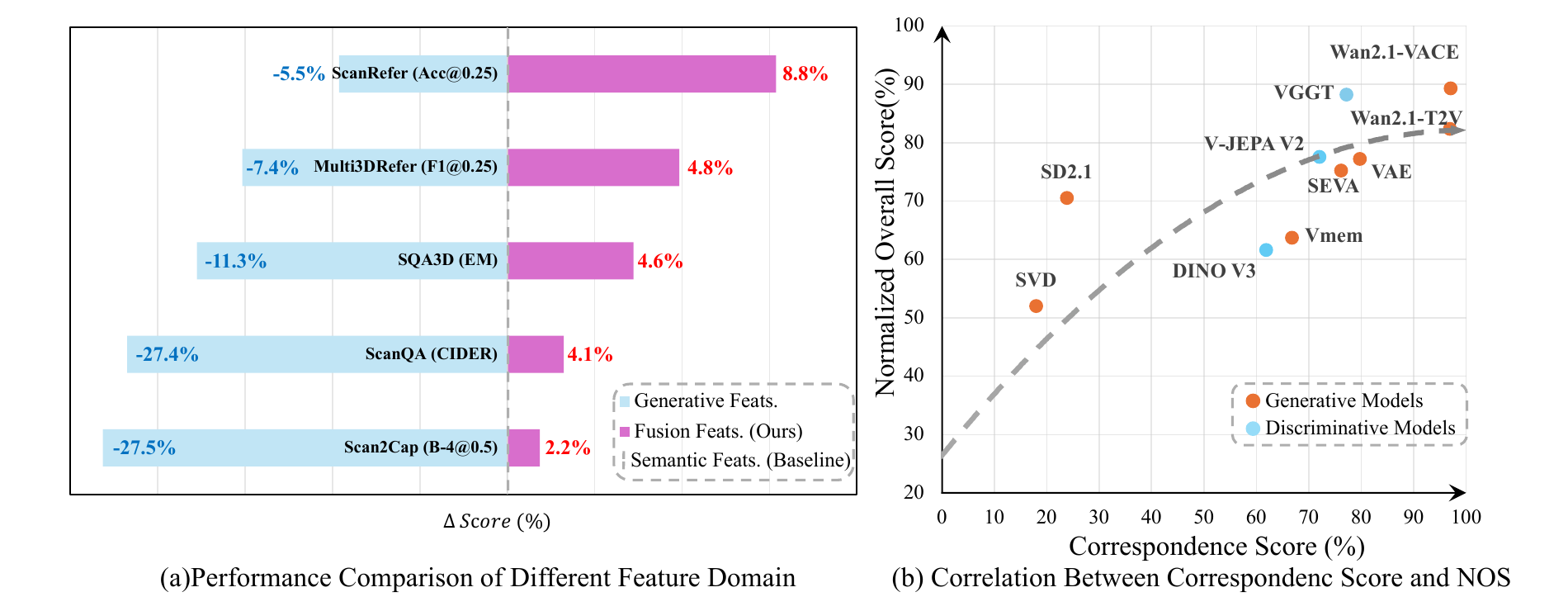}
\vspace{-10pt}
\caption{\textbf{Feature Analysis. (a):} Generative priors effectively complement semantic features with fusion, yielding consistent performance gains. \textbf{(b):} Multi-view correspondence strongly correlates with downstream 3D understanding performance. More details in Sec.~\ref{subsec: 3d_awareness}}
\vspace{-15pt}
\label{fig: domain&consistency}
\end{figure*}

Extensive experiments on 3D scene understanding (e.g., visual grounding, dense captioning, and QA), spatial reasoning benchmarks (e.g., VSI-Bench \cite{yang2025thinking}), and robotics manipulation tasks (LIBERO \cite{liu2024libero}) demonstrate that our method significantly outperforms larger spatially-enhanced models. Furthermore, in Fig.~\ref{fig: domain&consistency}, we provide quantitative evidence for the strong correlation between multi-view correspondence and downstream understanding performance. Besides, as evidenced in Fig.~\ref{fig: domain&consistency}(a), the gains stem from synergy rather than replacement: generative and semantic features are complementary, and their fusion yields substantial improvements. Our analysis further shows that the most informative spatial cues emerge from intermediate representations and mid-denoising time of the generative model, instead of the final pixel outputs, and that these priors are particularly beneficial for localization-centric tasks, effectively providing a spatial anchor for MLLMs.

In summary, our contributions are threefold.
\squishlist
    \item We investigate that modern video generators learn transferable spatiotemporal priors that encode geometry-consistent structure and motion, and we show that these priors are most informative in intermediate representations and mid-denoising stages.
    \item We propose \model{VEGA-3D}, a plug-and-play framework that repurposes video generation models as \textit{Latent World Simulator} for MLLMs, and introduce a token-level adaptive gated fusion module to align and integrate heterogeneous generative and semantic token spaces.
    \item Extensive experiments across 3D scene understanding, spatial reasoning, and embodied manipulation benchmarks demonstrate consistent gains, validating generative priors. Moreover, our framework is scalable: advances in video generation are readily transferable to stronger downstream 3D understanding.
\squishend
\section{Related Work}

\subsection{Scene Understanding with Large Language Models}

Extending Large Language Models to 3D domains is a rapidly growing field~\cite{deng2025best3dscenerepresentation}. Early approaches aligned point cloud encoders directly with LLMs \cite{pointllm, Point-bind, gpt4point, chatscene}, as seen in PointLLM \cite{pointllm}, Point-Bind \cite{Point-bind}, and GPT4Point \cite{gpt4point}. While effective, they heavily depend on the availability of high-quality 3D data. 
To bypass the need for direct 3D input, multi-view approaches \cite{video3dllm, wang2025ross3d, huang2025, gpt4scene, zhou2025llava} like Video-3D LLM \cite{video3dllm} and GPT4Scene \cite{gpt4scene} project 2D features into 3D space using positional embeddings or BEV rendering. More recent works attempt to lift 2D representations via auxiliary geometric supervision: Ross3D \cite{wang2025ross3d} utilizes reconstructive instruction tuning with bird's-eye-view supervision, while 3DRS \cite{huang2025} and ThinkWith3D \cite{chen2025think} distill knowledge from pre-trained 3D backbones. 
However, these methods typically require complex multi-stage training pipelines or task-specific geometric annotations (e.g., depth, camera pose). In contrast, our approach leverages the implicit physical priors already present in pre-trained video generation models, eliminating the need for explicit geometric supervision or complex rendering pipelines.

\subsection{Spatial Reasoning}

While MLLMs excel at semantic recognition, they often suffer from "spatial blindness" when tasked with geometric reasoning or determining spatial relationships, as highlighted by benchmarks \cite{jia2025omnispatial, zhang2025sphere, yang2025thinking, lin2025ost, yang2025mmsi} like Sphere \cite{zhang2025sphere} and VSI-Bench \cite{yang2025thinking}.
To mitigate this, one line of research focuses on scaling data: SpatialVLM \cite{chen2024spatialvlm} and VLM-3R \cite{fan2025vlm} train on massive datasets of spatial reasoning instructions to embed geometric concepts. Another direction explores mental simulation or chain-of-thought prompting~\cite{guan2026videostreamingthinking}, where models like MindCube \cite{yin2025spatial} and CVP \cite{chen2025cvp} verify spatial logic through auxiliary cognitive maps or reconstruction. 

Distinct from these approaches, which treat spatial reasoning as a linguistic or logical problem, we treat it as a representational problem. By fusing generative video priors, we ground the MLLM's reasoning in a physically consistent world model, enabling intuitive spatial understanding akin to human perception.

\subsection{Video Generation Models}

Video generation has rapidly progressed to high-fidelity, long-horizon synthesis \cite{videoworldsimulators2024, wan2025wan, kondratyuk2023videopoet, hong2022cogvideo, yang2024cogvideox}. Recent large-scale models demonstrate strong temporal coherence, suggesting their latent spaces capture rich spatiotemporal regularities \cite{videoworldsimulators2024, wan2025wan, kondratyuk2023videopoet}. Beyond visual fidelity, an emerging line of research focuses on structuring and controlling these generators for downstream applications \cite{genie3, li2025vmem, zhou2025stable, ren2025gen3c}.

Crucially, a recent wave of work connects video generation with 3D understanding, though typically by coupling generation with explicit geometric supervision. For instance, Ross3D~\cite{wang2025ross3d} and Omni-View~\cite{hu2025omniview} treat generation or novel-view synthesis as trainable auxiliary tasks, relying on explicit geometric labels (\eg, depth, camera poses, or bird's-eye-view targets) to inject constraints into the model. Other works integrate world models for reasoning and control: MILO~\cite{cao2025seeingimaginationlearningscene} utilizes implicit world modeling for spatial reasoning, GaussianDWM~\cite{Deng_2026_CVPR} couples a 3D Gaussian world model with unified scene understanding and generation, DyVA~\cite{zhang2025dyva} pairs a video model with a VLM to infer dynamics, and JEPA-VLA~\cite{miao2026jepavla} injects predictive embeddings into policies for robotic learning. In contrast, \model{VEGA-3D} keeps the video generator entirely \emph{frozen} and performs no generation or geometry estimation during training. We instead extract the multi-view feature consistency inherent in a frozen diffusion model as a label-free 3D prior. Our approach is thus orthogonal and complementary: advances in video generation translate directly into stronger 3D priors at no annotation cost.
\section{Preliminaries} \label{sec:pre}

\myparagraph{Multimodal Large Language Models.}
Following standard protocols~\cite{liu2023visual, radford2021learning}, we consider a multimodal large language model with parameters $\Theta$.
Given a multimodal input consisting of text tokens $\mathbf{x}$ and visual inputs $\mathbf{V}$, the visual content is mapped to a sequence of visual embeddings
$\mathbf{v}= f_{\text{proj}}\!\left(f_{\text{enc}}(\mathbf{V})\right)$,
where $f_{\text{enc}}$ is a visual encoder (e.g., SigLIP~\cite{zhai2023sigmoid}) and $f_{\text{proj}}$ is a projector.

The MLLM is trained to maximize the likelihood of the response token sequence $\mathbf{y}$ given the context:
\begin{equation}
\label{eq:lmm_loss}
\mathcal{L}_{\mathrm{CE}}(\Theta)
= - \sum_{i=1}^{L} \log p_{\Theta}\!\left(y_i \mid y_{<i}, \mathbf{x}, \mathbf{v}\right),
\end{equation}
where $\Theta$ denotes all trainable parameters (e.g., $\Theta=(\theta_{\text{lm}},\theta_{\text{enc}},\theta_{\text{proj}})$).

Crucially, this supervision is \textit{sparse} and \textit{discrete} \cite{assran2025v, chen2025vl}. The loss is computed in the vocabulary space, where spatial errors
(e.g., predicting ``left'' vs.\ ``right'') are treated as generic token mismatches. Lacking geometric metric constraints,
standard discriminative encoders $f_{\text{enc}}$ often exhibit ``spatial blindness,'' focusing on semantic presence rather than a precise spatial structure.

\myparagraph{Video Diffusion Models.}
Modern video generators (e.g., Wan2.1~\cite{wan2025wan}) are Diffusion Transformers trained with Flow Matching~\cite{lipman2022flow}, which learns a continuous-time transport field in the latent space.
Given a clean latent video $\mathbf{z}_0$, we sample Gaussian noise $\boldsymbol{\epsilon}\sim\mathcal{N}(\mathbf{0},\mathbf{I})$ and a time $t\sim\mathcal{U}(0,1)$, and train a flow network $v_{\psi}(\cdot)$ to regress the target velocity under MSE:
\begin{equation}
\label{eq:gen_loss}
\mathcal{L}_{\mathrm{FM}}(\psi)
= \mathbb{E}_{\mathbf{z}_0,\boldsymbol{\epsilon},t}\!\left[
\left\| \mathbf{u}_t - v_{\psi}(\mathbf{z}_t,t,\mathbf{c}) \right\|_2^2
\right],
\end{equation}
where $\mathbf{c}$ denotes conditioning signals.
The corresponding target velocity is
$\mathbf{u}_t = \frac{\mathrm{d}\mathbf{z}_t}{\mathrm{d}t}$.
In implementation, we use a discrete timestep index $k\in\{0,\ldots,K\}$ (with $K{=}1000$) and its normalized time $t_k=\frac{k}{K}$.

\section{Method}
\label{sec:method}

\begin{figure*}[!t]
\centering
\includegraphics[width=\textwidth]{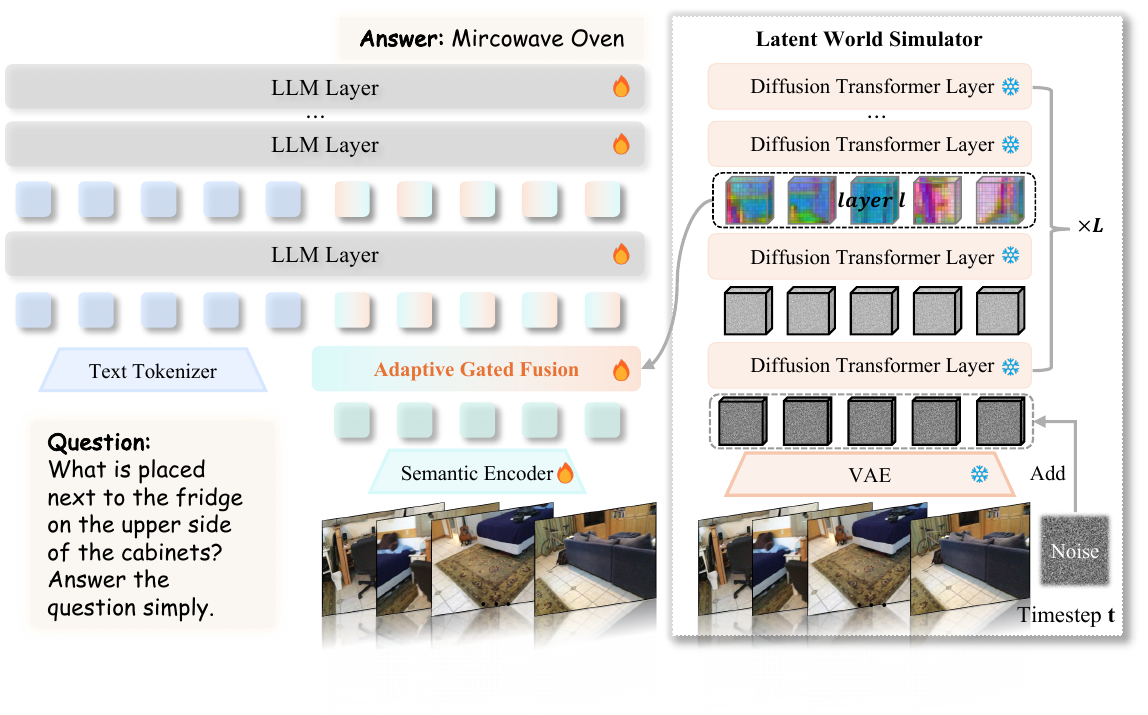}
\vspace{-30pt}
\caption{\textbf{Overview of the VEGA-3D framework.} We repurpose a frozen video generation model as a \textbf{Latent World Simulator} to extract implicit 3D priors. These features are dynamically integrated with the semantic stream via \textbf{Adaptive Gated Fusion}, equipping the MLLM with dense 3D structural awareness.}
\label{fig: model}
\vspace{-3pt}
\end{figure*}

Our goal is to endow Multimodal Large Language Models (MLLMs) with the implicit generative prior inherent in video generation models.
As illustrated in \cref{fig: model}, our framework introduces a dual-branch visual encoding mechanism that synergizes the high-level semantic capabilities of a discriminative encoder (e.g., SigLIP~\cite{zhai2023sigmoid}) and dense 3D structure priors from a generative video diffusion model (e.g., Wan2.1~\cite{wan2025wan}).
The methodology is organized into three logical stages:
(1) \textbf{3D Awareness Analysis} (Sec.~\ref{subsec: 3d_awareness}), where we identify multi-view feature consistency as the key indicator of geometric capability;
(2) \textbf{Latent World Simulation} (Sec.~\ref{subsec: gen_extract}), which operationalizes these insights by mining spatiotemporal geometry from the generator's intermediate representations via noise injection; and
(3) \textbf{Bridging the Generative and Semantic Gap} (Sec.~\ref{subsec: fusion}), which adaptively integrates these heterogeneous features via a token-level adaptive gated fusion mechanism to align with the MLLM.

\subsection{3D Awareness via Multi-view Feature Consistency}
\label{subsec: 3d_awareness}

A pivotal factor in robust 3D scene understanding is the ability to maintain consistent representations of physical geometry across varying viewpoints. While traditional discriminative models excel at semantic invariance, we hypothesize that effective 3D reasoning often benefits from multi-view feature consistency, which maps the same physical 3D point to a unified latent representation across different views. To quantitatively verify this correlation and evaluate the geometric integrity of different feature backbones, we introduce \textbf{Multi-view Correspondence Score}.

\myparagraph{Metric Definition.}
We utilize the ScanNet test dataset split \cite{scannet}, which provides posed RGB frames and dense depth maps (only for analysis). For a 3D scene observed from $V$ views, we project the encoder features $\mathbf{F}_v$ from each view into a shared global voxel grid using the ground-truth camera extrinsics and depth.
For a specific voxel $m$ observed in two different views $v_i$ and $v_j$, we extract the corresponding feature vectors $\mathbf{h}_{m, v_i}$ and $\mathbf{h}_{m, v_j}$. The consistency score for this voxel is defined as cosine similarity:
\begin{equation}
S_{\text{voxel}}^{(m)} = \frac{\mathbf{h}_{m, v_i}^\top \mathbf{h}_{m, v_j}}{\|\mathbf{h}_{m, v_i}\| \|\mathbf{h}_{m, v_j}\|}.
\end{equation}
The final scene-level score is obtained by averaging $S_{\text{voxel}}^{(m)}$ over all valid voxel pairs across the scene. A higher score indicates that the model implicitly aligns distinct views of the same 3D structure.

\myparagraph{Correlation and Architectural Analysis.}
To validate whether this consistency serves as a reliable indicator for downstream 3D capability, we define a Normalized Overall Score \textbf{(NOS)}. It is calculated by normalizing the performance metrics in Tab.~\ref{tab:video_models} to $[0, 1]$ with Min-Max normalization across the discriminative and generative models, explicitly including the baseline results to establish a relative performance improvement, and then averaging them into a single scalar.

As illustrated in Fig.~\ref{fig: domain&consistency}, plotting the Correspondence Score against NOS reveals a distinct positive correlation, confirming that multi-view consistency is a strong predictor of 3D performance.
Furthermore, the results highlight a significant architectural divergence. Models based on UNet architectures (e.g., SVD \cite{blattmann2023stable}, Stable Diffusion \cite{rombach2022high}, Vmem \cite{li2025vmem}) exhibit notably lower consistency scores. We attribute this to the local inductive bias of convolutions and the insufficient scale of data, which limits the receptive field and hinders long-range geometric alignment. In contrast, DiT based models (e.g., Wan2.1 \cite{wan2025wan}) leverage global attention mechanisms to capture holistic context, achieving remarkably high consistency ($>96\%$) and consequently superior downstream 3D understanding.

Guided by this evidence, \model{VEGA-3D} selects DiT-based architectures to provide robust spatial priors. Next, we detail how to actively extract these implicit geometric representations from a frozen generative model.

\subsection{Video Generative Model as a Latent World Simulator}
\label{subsec: gen_extract}

\begin{wrapfigure}{r}{0.4\textwidth}
  \centering
  \vspace{-25pt}
  \includegraphics[width=\linewidth]{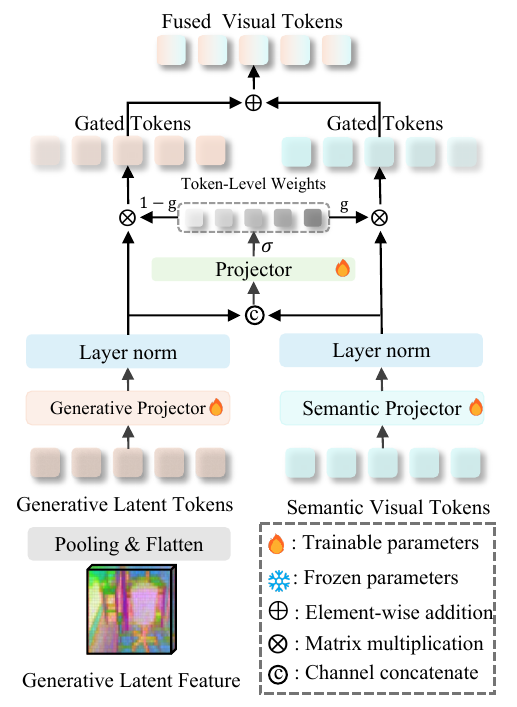} 
  \caption{\textbf{Adaptive Gated Fusion.} It dynamically integrates heterogeneous features using a token-level gating mechanism.}
  \vspace{-20pt}
  \label{fig:fusion}
\end{wrapfigure}

Building on the premise that generative models encapsulate physical laws, we adopt the pretrained, parameter-frozen Wan2.1-T2V 1.3B~\cite{wan2025wan} as our default generative encoder for its simple text-conditioning interface and strong localization-centric performance. We additionally evaluate other video generative models in Tab.~\ref{tab:video_models}, demonstrating that our framework is compatible with different video generative backbones. While traditional visual encoders process raw pixel intensities, video generative models operate in a compressed latent space governed by diffusion dynamics.

Given an input video sequence $\mathbf{V}\in\mathbb{R}^{T\times H\times W\times 3}$ of $T$ frames, we first map it to a low-dimensional latent space via the model's Variational Autoencoder (VAE), yielding $\mathbf{z}_0 = E(\mathbf{V})$.
However, a static latent representation $\mathbf{z}_0$ is insufficient to activate the generative model's reasoning capabilities fully.
Diffusion models are trained to enforce structural coherence primarily during active denoising of a corrupted signal; the process of restoration reveals the model's understanding of structure.
Therefore, we perturb the clean latent along the same Flow Matching noising path used by the pretrained backbone.
Specifically, we choose a discrete timestep index $k\in\{0,\ldots,K\}$ and define the normalized time as $t_k=\frac{k}{K}$.
We then sample $\boldsymbol{\epsilon}\sim\mathcal{N}(\mathbf{0},\mathbf{I})$ and construct:
\begin{equation}
\label{eq:forward_noising}
\mathbf{z}_{k} = (1-t_k)\,\mathbf{z}_0 + t_k\,\boldsymbol{\epsilon}.
\end{equation}

We feed $\mathbf{z}_k$ into the backbone $\Phi(\cdot)$ using an empty text prompt ($\mathbf{c}_{\text{text}}=\texttt{""}$). This ensures that the activated features rely solely on the visual signal and the model's learned physics, minimizing semantic hallucination.
We empirically select features from the specific intermediate DiT layer $l$, as they offer an optimal trade-off between spatial precision and abstract spatiotemporal context:
\begin{equation}
\label{eq:gen_feat_raw}
\mathbf{f}_{\mathrm{raw}} = \Phi^{(l)}(\mathbf{z}_k, k; \mathbf{c}_{\text{text}}=\texttt{""}).
\end{equation}
After Adaptive Average Pooling to match the semantic tokenization, we obtain the generative representation $\mathbf{f}_{\mathrm{gen}}\in\mathbb{R}^{T\times N\times D_{\mathrm{gen}}}$. While this noise-driven process effectively extracts implicit 3D knowledge, these continuous physical features inherently misalign with the MLLM's discrete semantic space. To bridge this semantic-geometric gap, \model{VEGA-3D} introduces a tailored fusion strategy.

\subsection{Bridging the Generative and Semantic Gap}
\label{subsec: fusion}

The generative features $\mathbf{f}_{\mathrm{gen}}$ and semantic features $\mathbf{f}_{\mathrm{sem}}$ reside in fundamentally different manifolds. To effectively synergize them, we introduce a mechanism to bridge this gap.

As shown in Fig.~\ref{fig:fusion}, we first project both streams into the LLM's hidden dimension $D_{\mathrm{llm}}$ via independent MLP projectors $P_{\mathrm{gen}}$ and $P_{\mathrm{sem}}$, aligning them to a shared embedding space:
\begin{equation}
\label{eq:proj}
\mathbf{F}_{\mathrm{gen}} = P_{\mathrm{gen}}(\mathbf{f}_{\mathrm{gen}}),
\mathbf{F}_{\mathrm{sem}} = P_{\mathrm{sem}}(\mathbf{f}_{\mathrm{sem}}).
\end{equation}
Here, $\mathbf{F}_{\mathrm{gen}}, \mathbf{F}_{\mathrm{sem}} \in \mathbb{R}^{T \times N \times D_{\mathrm{llm}}}$, where $T$ is the number of frames and $N$ is the number of tokens per frame.

To avoid simply averaging conflicting signals, we employ an \textbf{Adaptive Gated Fusion} mechanism. This allows the model to adaptively weigh semantic versus structural cues for each specific token location. For the $i$-th spatial token $\mathbf{F}_i$, we compute a scalar gate $g_i \in [0,1]$:
\begin{equation}
\label{eq:gate}
g_i
=
\sigma\!\left(
\mathbf{W}_g^{\top} \,
\mathrm{Concat}\left(
\mathrm{LN}(\mathbf{F}_{\mathrm{gen}, i}), \,
\mathrm{LN}(\mathbf{F}_{\mathrm{sem}, i})
\right)
+ b_g
\right),
\end{equation}
where $\sigma(\cdot)$ is the sigmoid function, $\mathrm{LN}$ denotes Layer Normalization, and $\mathbf{W}_g$ is a learnable weight vector. The final fused representation is a convex combination determined by this gate:
\begin{equation}
\label{eq:fused}
\mathbf{F}^{\mathrm{fused}}_i
=
(1 - g_i) \cdot \mathbf{F}_{\mathrm{gen}, i}
+
g_i \cdot \mathbf{F}_{\mathrm{sem}, i}.
\end{equation}

Crucially, this gate $g_i$ acts as a semantic-geometric arbitrator: it enables the model to prioritize semantic priors for recognition tasks, while dynamically shifting attention to generative world knowledge for tasks requiring spatial reasoning. By seamlessly integrating continuous spatial priors with discrete semantic representations, \model{VEGA-3D} overcomes the spatial blindness of traditional encoders, achieving dense 3D understanding without explicit geometric supervision.
\label{subsec:obj}

\section{Experiments}

\begin{table*}[!t]
\centering
\scriptsize
\setlength{\tabcolsep}{0.4mm}
\caption{Performance comparison on 3D scene understanding benchmarks. Specialists are single-task methods, while generalists target multiple tasks. $\dagger$ indicates the model is trained on extra datasets; $\ddagger$ indicates the model additionally uses BEV (bird's-eye-view) reconstruction supervision rendered from depth and camera poses. \textbf{Bold} marks the best result in each column, and the Avg. Rank is recomputed over all listed methods. The baseline model is Video-3D LLM.}
\begin{tabular}{llccccccccc|c}
\toprule
\multirow{2}{*}{Method} & \multirow{2}{*}{Ref.}
& \multicolumn{2}{c}{ScanRefer} 
& \multicolumn{2}{c}{Multi3DRefer} 
& \multicolumn{2}{c}{Scan2Cap}  
& \multicolumn{2}{c}{ScanQA} 
& SQA3D & Avg. \\ 
\cmidrule(lr){3-4} \cmidrule(lr){5-6} \cmidrule(lr){7-8} \cmidrule(lr){9-10} \cmidrule(lr){11-11}
&  & Acc$_{.25}$ & Acc$_{.5}$ & F1$_{.25}$ & F1$_{.5}$ & C$_{.5}$ & B-4$_{.5}$ & C & EM & EM & Rank \\
\midrule
\multicolumn{12}{l}{\textit{\textbf{Specialists}}} \\
ScanRefer~\cite{scanrefer}      & ECCV 20    & 37.3 & 24.3 & --   & --   & --   & --   & --   & --   & --   & 18.5 \\
MVT~\cite{mvt}                  & CVPR 22    & 40.8 & 33.3 & --   & --   & --   & --   & --   & --   & --   & 17.5 \\
3DVG-Trans~\cite{3dvg-trans}    & ICCV 21    & 45.9 & 34.5 & --   & --   & --   & --   & --   & --   & --   & 16.0 \\
ViL3DRel~\cite{vil3drel}        & NeurIPS 21 & 47.9 & 37.7 & --   & --   & --   & --   & --   & --   & --   & 14.5 \\
M3DRef-CLIP~\cite{multi3drefer} & ICCV 23    & 51.9 & 44.7 & 42.8 & --   & 38.4 & --   & --   & --   & --   & 11.5 \\
Scan2Cap~\cite{scan2cap}        & CVPR 21    & --   & --   & --   & --   & 35.2 & 22.4 & --   & --   & --   & 17.5 \\
ScanQA~\cite{scanqa}            & CVPR 22    & --   & --   & --   & --   & --   & --   & 64.9 & 21.1 & 47.2 & 14.3 \\
3D-VisTA~\cite{3dvista}         & ICCV 23    & 50.6 & 45.8 & --   & --   & 66.9 & 34.0 & 69.6 & 22.4 & 48.5 & 12.4 \\
\midrule
\multicolumn{12}{l}{\textit{\textbf{Generalists}}} \\
Chat-3D~\cite{chat3d}                     & Arxiv 23   & --   & --   & --   & --   & --   & --   & 53.2 & --   & --   & 18.0 \\
Chat-3D v2~\cite{chatscene}               & Arxiv 23   & 42.5 & 38.4 & 45.1 & 41.6 & 63.9 & 31.8 & 87.6 & --   & 54.7 & 12.5 \\
LL3DA~\cite{ll3da}                        & CVPR 24    & --   & --   & --   & --   & 62.9 & 36.0 & 76.8 & --   & --   & 14.0 \\
LEO~\cite{leo}                            & ICML 24    & --   & --   & --   & --   & 72.4 & 38.2 & 101.4 & 21.5 & 50.0 & 10.0 \\
Grounded3D-LLM~\cite{grounded-3dllm}      & Arxiv 24   & 47.9 & 44.1 & 45.2 & 40.6 & 70.6 & 35.5 & 72.7 & --   & --   & 12.3 \\
PQ3D~\cite{pq3d}                          & ECCV 24    & 57.0 & 51.2 & --   & 50.1 & 80.3 & 36.0 & --   & --   & 47.1 & 9.2 \\
ChatScene~\cite{chatscene}                & NeurIPS 24 & 55.5 & 50.2 & 57.1 & 52.4 & 77.1 & 36.3 & 87.7 & 21.6 & 54.6 & 8.8 \\
SceneLLM~\cite{scenellm}                  & WACV 25    & --   & --   & --   & --   & --   & --   & 80.0 & 27.2 & 53.6 & 10.3 \\
Inst3D-LLM~\cite{inst3d}                  & CVPR 25    & 57.8 & 51.6 & 58.3 & 53.5 & 79.7 & 38.3 & 88.6 & 24.6 & --   & 6.8 \\
3D-LLaVA~\cite{3dllava}                   & CVPR 25    & 51.2 & 40.6 & --   & --   & 78.8 & 36.9 & 92.6 & --   & 54.5 & 9.8 \\
3DRS~\cite{huang2025}                     & NeurIPS 25 & 62.9 & 56.1 & 60.4 & 54.9 & \textbf{86.1} & 41.6 & 104.8 & 30.3 & 60.6 & 2.7 \\
LLaVA-3D~\cite{llava3d}                   & ICCV 25    & 50.1 & 42.7 & 49.8 & 43.6 & 84.1 & 42.6 & --   & 30.6 & 60.1 & 6.6 \\
Ross3D$\ddagger$~\cite{wang2025ross3d}    & ICCV 25    & 61.1 & 54.4 & 59.6 & 54.3 & 81.3 & 43.4 & \textbf{107.0} & \textbf{30.8} & \textbf{63.0} & 2.6 \\
LLaVA-4D$\dagger$~\cite{zhou2025llava}    & ICLR 26    & --   & 53.2 & --   & 54.3 & 85.3 & \textbf{45.7} & 97.8 & --   & --   & 3.4 \\
Omni-View$\dagger$~\cite{hu2025omniview}  & ICLR 26    & 50.8 & 45.0 & --   & --   & --   & --   & 103.0 & 29.5 & 59.2 & 7.0 \\
Fase3D~\cite{mei2026efficientencoderfreefourierbased3d} & CVPR 26 & -- & -- & -- & -- & 78.1 & 41.3 & 91.7 & -- & 54.3 & 8.8 \\
\midrule
\rowcolor{baselinegray}
Video-3D LLM~\cite{video3dllm}            & CVPR 25    & 58.1 & 51.7 & 58.0 & 52.7 & 83.8 & 41.3 & 102.1 & 30.1 & 58.6 & 5.1 \\
\textbf{VEGA-3D (Ours)}                   & -          & \textbf{63.2} & \textbf{56.2} & \textbf{60.8} & \textbf{55.1} & 83.2 & 42.2 & 106.3 & 30.4 & 61.3 & \textbf{2.2} \\
\bottomrule
\end{tabular}
\vspace{-15pt}
\label{tab:scene_understanding}
\end{table*}

\subsection{Implementation Details}

We evaluate our framework on three representative axes: (i) \textbf{3D scene understanding} on ScanRefer \cite{scanrefer}, Multi3DRefer \cite{multi3drefer}, Scan2Cap \cite{scan2cap}, ScanQA \cite{scanqa}, and SQA3D \cite{sqa3d}; (ii) \textbf{spatial reasoning} on VSI-Bench \cite{yang2025thinking} with diverse capability categories and (iii) \textbf{robotic manipulation} on LIBERO \cite{liu2024libero} with four task suites and their average success rate. These benchmarks and reported metrics follow the standard protocols summarized in our main tables.

For 3D scene understanding, we build upon Video-3D LLM \cite{video3dllm} as our baseline generalist and select Wan2.1-T2V 1.3B \cite{wan2025wan} as the latent world simulator plus an adaptive gated fusion module. For VSI-Bench \cite{yang2025thinking}, we adopt Qwen2.5VL-7B \cite{Qwen2.5-VL} as the baseline and attach the same plug-and-play generative branch, and the training datasets follow VG-LLM \cite{vg-llm}. For LIBERO \cite{liu2024libero}, we start from OpenVLA-OFT \cite{kim2025fine} and inject generative priors into the visual stream before policy learning. This design keeps the overall training and evaluation pipelines consistent with the corresponding baselines, while isolating the effect of generative priors. More details are provided in the Appendix.

For both training and inference, we uniformly sample 32 frames per scan to construct multi-view image sets. The Flow-Matching time interval $t\in[0,1]$ is discretized into $K{=}1000$ steps in the pretrained Wan2.1 backbone. We denote the discrete timestep index as $k$ and use $t_k=\frac{k}{K}$ as the normalized time. By default, we extract features at $k{=}300$ (i.e., $t_k{=}0.3$) from the 20th DiT layer. When calculating the correspondence score, we use a voxel size of 0.1 for voxelization. All models are optimized using Adam, with a batch size of 128 and a warm-up ratio of 0.03. The learning rates are set to a maximum of $1\times10^{-5}$ for the language model and $2\times10^{-6}$ for the visual backbone during the warm-up period. We use 8 H100 NVIDIA GPUs for all experiments.

\begin{table*}[!t]
\vspace{-2.5mm}
    \centering
    \scriptsize
    \setlength{\tabcolsep}{0.8mm}
    \caption{{The comparison with state-of-the-art models on VSI-Bench.} \textit{Spatial-Enhanced Models} are models that are specialized for spatial reasoning. $\dagger$ indicates the baseline model's performance is finetuned on the same training dataset configurations.
    The baseline model is Qwen2.5VL-7B. }
    \vspace{-10pt}
    \begin{tabular}{l|c|cccccccc}
    & & 
    \rotatebox{60}{{Obj. Count}} &
    \rotatebox{60}{{Abs. Dist.}} &
    \rotatebox{60}{{Obj. Size}} & 
    \rotatebox{60}{{Room Size}} &
    \rotatebox{60}{{Rel. Dist.}} &
    \rotatebox{60}{{Rel. Dir.}} &
    \rotatebox{60}{{Route Plan}} &
    \rotatebox{60}{{Appr. Order}} \\
    \makecell[c]{{Model}} & {Avg.} & \multicolumn{4}{c}{\cellcolor{orange!10}{Numerical Answer}} & \multicolumn{4}{c}{\cellcolor{yellow!10}{Multiple-Choice Answer}} \\
    \hline
    \rowcolor{navyblue!5}
    \hline
    
    \rowcolor{navyblue!5}
    \multicolumn{1}{l|}{\textcolor{black}{\textit{Proprietary Models (API)}}} & & & & & & & & & \\
    GPT-4o~\cite{hurst2024gpt4o} & 34.0 & 46.2 & 5.3 & 43.8 & 38.2 & 37.0 & 41.3 & 31.5 & 28.5 \\
    Gemini-1.5-Pro~\cite{team2024gemini} & 45.4 & 56.2 & 30.9 & 64.1 & 43.6 & 51.3 & 46.3 & 36.0 & 34.6 \\
    Gemini-1.5-Flash~\cite{team2024gemini} & 42.1 & 49.8 & 30.8 & 53.5 & {54.4} & 37.7 & 41.0 & 31.5 & 37.8 \\
    \hline
    \rowcolor{navyblue!5}
    \multicolumn{1}{l|}{\textcolor{black}{\textit{Open-source Models}}} & & & & & & & & & \\
    LongVA-7B~\cite{zhang2024long} & 29.2 & 38.0 & 16.6 & 38.9 & 22.2 & 33.1 & 43.3 & 25.4 & 15.7 \\
    LongVILA-8B~\cite{chen2024longvila}  & 21.6 & 29.1 & 9.1 & 16.7 & 0.0 & 29.6 & 30.7 & 32.5 & 25.5 \\
    InternVL2-8B~\cite{chen2024internvl2} & 34.6 & 23.1 & {28.7} & 48.2 & {39.8} & 36.7 & 30.7 & 29.9 & 39.6 \\
    InternVL2-40B~\cite{chen2024internvl2} & 36.0 & 34.9 & 26.9 & 46.5 & 31.8 & 42.1 & 32.2 & 34.0 & 39.6 \\
    VILA-1.5-40B~\cite{liu2025nvila}  & 31.2 & 22.4 & 24.8 & 48.7 & 22.7 & 40.5 & 25.7 & 31.5 & 32.9 \\
    LLaVA-OneVision-7B~\cite{li2024llava-onevision}  & 32.4 & 47.7 & 20.2 & 47.4 & 12.3 & 42.5 & 35.2 & 29.4 & 24.4 \\
    LLaVA-OneVision-72B~\cite{li2024llava-onevision} & 40.2 & 43.5 & 23.9 & {57.6} & 37.5 & 42.5 & 39.9 & 32.5 & 44.6 \\
    LLaVA-NeXT-Video-7B~\cite{liu2024llavanext} & 35.6 & 48.5 & 14.0 & 47.8 & 24.2 & {43.5} & 42.4 & 34.0 & 30.6 \\
    LLaVA-NeXT-Video-72B~\cite{liu2024llavanext}  & 40.9 & {48.9} & 22.8 & 57.4 & 35.3 & 42.4 & 36.7 & {35.0} & {48.6} \\
    \hline
    \rowcolor{navyblue!5}
    \multicolumn{1}{l|}{\textcolor{black}{\textit{Spatial-Enhanced Models}}} & & & & & & & & & \\
    Video-R1-7B~\cite{feng2025video} & 37.1 & - & - & - & - & - & - & -  & - \\
    vsGRPO-V-7B~\cite{liao2025improved} & 40.7 & 59.9 & 29.6 & 50.8 & 48.3 & 35.4 & 35.6 & 34.0  & 31.5 \\
    SPAR-8B~\cite{zhang2025flatland} & 41.1 & - & - & - & - & - & - & -  & - \\
    SpaceR-7B~\cite{ouyang2025spacer} & 45.6 & - & - & - & - & - & - & -  & - \\
    3DRS-7B~\cite{huang2025} & 45.9 & 68.7 & 34.8 & 53.6 & 56.6 & 40.9 & 43.2 & 30.4  & 39.2 \\
    VG-LLM-4B~\cite{zheng2025learning} & 45.9 & 65.6 & 37.4 & 54.8 & 60.2 & 42.3 & 46.3 & 33.0 & 25.9 \\
    VG-LLM-8B~\cite{zheng2025learning} & 50.1 & 67.2 & 38.0 & 59.3 & 63.2 & 47.0 & 43.9 & 33.0 & 49.4 \\
    \hline
    \rowcolor{baselinegray}
    Qwen2.5VL-7B $\dagger$ ~\cite{Qwen2.5-VL} & 48.9 & 68.3 & 37.0 & 57.4 & 58.7 & 39.7 & 43.0 & 29.4  & 57.8 \\
    \textbf{VEGA-3D (Ours)} & \textbf{50.5} & 69.7 & 35.9 & 58.0 & 60.8 & 45.1 & 43.1 & 30.9 & 60.5 \\
    \hline
    \end{tabular}
\label{tab:vsi_bench}
\vspace{-15pt}
\end{table*}

\subsection{Main Results on 3D Scene Understanding}

Tab.~\ref{tab:scene_understanding} reports the main results on five 3D scene understanding benchmarks, covering spatial grounding, dense captioning, and question answering. Overall, our VEGA-3D consistently improves over the Video-3D LLM \cite{video3dllm} baseline across most metrics, particularly excelling in localization-centric tasks (e.g., boosting ScanRefer Acc@0.5 from 51.7 to 56.2, SQA3D EM 58.6 to 61.3).

This performance divergence across different tasks suggest an informative pattern about when generative priors help most. We observe notable gains in grounding and spatial QA, where the implicit 3D structural awareness extracted from the generative backbone acts as a robust spatial anchor, which helps reduce the spatial ambiguity of standard MLLMs (see Fig.~\ref{fig: consistency&attn}(b)). The slight CIDEr drop in Scan2Cap may indicate a semantic-geometry trade-off: emphasizing structural cues may weaken fine-grained lexical details. Our Adaptive Gated Fusion aims to balance this by token-wise weighting between the two streams.

Notably, in contrast to prevailing state-of-the-art methods that rely on heavy geometric supervision \cite{huang2025} via external 3D teachers \cite{wang2025vggt}, BEV reconstruction targets rendered from depth and camera poses \cite{wang2025ross3d}, or curated 3D-heavy datasets \cite{zhou2025llava}, our improvements are achieved entirely without explicit 3D annotations. In particular, while the reconstruction-based Ross3D~\cite{wang2025ross3d} attains higher ScanQA and SQA3D scores through its BEV supervision, \model{VEGA-3D} still achieves the best overall average rank and leads on every grounding-oriented metric, suggesting that label-free generative priors are competitive with explicit geometric supervision for localization-centric understanding. This conveys a powerful insight: large-scale video generation models, through the pretext of temporal synthesis, have already internalized a robust 3D world model from the vast causality of the natural world. By repurposing these models as Latent World Simulators, we bypass the data-scarcity bottleneck posed by 3D-specific labels. This framework offers a highly scalable, data-efficient paradigm, demonstrating that the next frontier for 3D spatial awareness in MLLMs may not lie in more 3D data but in unleashing the physical priors already dormant within generative foundations.

\vspace{-10pt}
\subsection{Generalization to Spatial Reasoning and Manipulation}

To validate the generalization ability of VEGA-3D, we extend our evaluation to 3D visual-spatial reasoning and embodied manipulation. 

\myparagraph{Spatial reasoning on VSI-Bench.} Tab.~\ref{tab:vsi_bench} evaluates spatial reasoning on VSI-Bench \cite{yang2025thinking}, a comprehensive benchmark designed to diagnose diverse visual-spatial skills from videos, such as relative distance and route planning. By seamlessly augmenting the Qwen2.5VL-7B \cite{Qwen2.5-VL} baseline with our generative priors, we observe consistent gains across the overall average and multiple sub-categories. This trend aligns with the recent emphasis on geometry-aware mechanisms for spatial reasoning, yet our method remains lightweight and plug-and-play.

\begin{table}[htbp] 
    \centering
    \vspace{-15pt}
    \caption{The comparison of the simulation robotic manipulation benchmark LIBERO, the performance is evaluated with the average success rate SR (\%).}
    \label{tab: libero}
    \scriptsize 
    \setlength{\tabcolsep}{1.2mm} 
    \renewcommand{\arraystretch}{1.1} 
    \begin{tabular*}{\linewidth}{@{\extracolsep{\fill}} lcccccc}
        \toprule
        Method & Reference & Spatial & Object & Goal & Long & Avg. \\
        \midrule
        Diffusion Policy \cite{chi2023diffusion} & ICRR 23 & 78.3 & 92.5 & 68.3 & 50.5 & 72.4 \\
        Octo \cite{team2024octo} & RSS 24 & 78.9 & 85.7 & 84.6 & 51.1 & 75.1 \\
        OpenVLA \cite{kim2024openvla} & CoRL 24 & 84.7 & 88.4 & 79.2 & 53.7 & 76.5 \\
        DiT Policy \cite{hou2025dita} & ICCV 25 & 84.2 & 96.3 & 85.4 & 63.8 & 82.4 \\
        CoT-VLA \cite{CoT-VLA-2025} & CVPR 25 & 87.5 & 91.6 & 87.6 & 69.0 & 81.1 \\
        UniVLA \cite{UniVLA} & RSS 25 & 96.5  & 96.8  & 95.6  & 92.0 & 95.2 \\
        \rowcolor{baselinegray} 
        OpenVLA-OFT \cite{kim2025fine} & RSS 25 & \textbf{97.5} & \underline{98.3} & \textbf{97.8} & \underline{94.4} & \underline{97.0} \\
        \textbf{VEGA-3D (Ours)}  & - & \underline{97.4} & \textbf{99.4} & \underline{97.0} & \textbf{95.2} & \textbf{97.3} \\
        \bottomrule
    \end{tabular*}
    \vspace{-20pt}
\end{table}

\myparagraph{Robotic manipulation on LIBERO.} Beyond passive reasoning, we assess whether our generative priors can ground active physical agents. Tab.~\ref{tab: libero} reports success rates on the LIBERO \cite{liu2024libero} suite, a challenging simulation benchmark for robotic manipulation. We treat \model{VEGA-3D} as a drop-in visual enhancement for Vision-Language-Action (VLA) models: starting from a pre-trained OpenVLA-OFT \cite{kim2025fine}, we leave the language backbone and action head unchanged and only augment the visual stream, injecting the frozen Latent World Simulator features through the same adaptive gated fusion module used for scene understanding before policy learning. Despite being extracted without explicit action-conditioned training, the generative priors improve the already highly saturated baseline, with the clearest gains on the object-centric and long-horizon suites that demand precise contact localization and persistent spatial memory. This indicates that the spatial regularities embedded in the Latent World Simulator transfer to action grounding, supplying the policy with a geometry-aware visual representation rather than additional action supervision.

\myparagraph{Towards World-Action Models.} These results suggest a shift towards World-Action Models (WAMs): instead of relying on explicit 3D encoders, a frozen video generator acts as a foundational World Model to directly ground the action policy. This synergy offers three key advantages: \textbf{(i) Modularity:} Spatial priors are injected at the visual-token level, leaving the policy and language backbones intact. \textbf{(ii) Scalability:} Physical intelligence stems from large-scale video pre-training, meaning advances in world simulation directly translate to stronger action grounding without extra robotic annotations.

\myparagraph{Real-robot manipulation.} To test transfer beyond simulation, we deploy our model on real Agilex Piper arms following the RoboTwin \cite{mu2025robotwin} protocol. On the Click Bell task, VEGA-3D improves over the baseline policy from 40\% (8/20) to 55\% (11/20), indicating that the prior can transfer to real-world manipulation.

\subsection{Ablation Studies}
\label{sec:ablation}

To validate our design choices, we conduct comprehensive ablation studies using Wan2.1-T2V as the default generative encoder. We note that VACE can achieve higher QA scores, while T2V provides stronger grounding-oriented performance; thus, we keep T2V as the default encoder.

\myparagraph{Generative vs. Discriminative Priors.} 

\noindent Tab.~\ref{tab:video_models} and Fig.~\ref{fig: domain&consistency}(b) reveal a strong positive correlation between multi-view consistency and 3D scene understanding. While traditional discriminative models (e.g., DINO V3 \cite{simeoni2025dinov3}, V-JEPA v2 \cite{assran2025v}, SigLIP \cite{zhai2023sigmoid}) offer rich semantics, they lack explicit 3D consistency. Conversely, DiT-based generative models and 3D foudation model like VGGT \cite{wang2025vggt} excel at capturing robust spatial priors. DiTs significantly outperform UNet-based models, as their global attention mechanisms preserve long-range geometric dependencies better than local convolutions. This confirms that video generation models provide superior 3D prior for spatial reasoning compared to standard visual learners.

\begin{table}[!t]
    \centering
    \scriptsize
    \setlength{\tabcolsep}{0.8mm}
    \caption{Experiments on using different discriminative and generative foundation models. Bold denotes the best in each group.}
    \vspace{-5pt}
    \label{tab:video_models}
        \begin{tabular}{llccccccccc}
            \toprule
            \multirow{2}{*}{Models}
            & \multirow{2}{*}{Params}
            & \multicolumn{2}{c}{ScanRefer} 
            & \multicolumn{2}{c}{Multi3DRefer} 
            & \multicolumn{2}{c}{Scan2Cap}  
            & \multicolumn{2}{c}{ScanQA} 
            & SQA3D \\
            \cmidrule(lr){3-4} \cmidrule(lr){5-6} \cmidrule(lr){7-8} \cmidrule(lr){9-10} \cmidrule(lr){11-11}
            & & Acc$_{.25}$ & Acc$_{.5}$ & F1$_{.25}$ & F1$_{.5}$ & C$_{.5}$ & B-4$_{.5}$ & C & EM & EM \\
            \midrule
            
            \rowcolor{baselinegray} 
            Baseline & - & 58.1 & 51.7 & 58.0 & 52.7 & 83.8 & 41.3 & 102.1 & 30.1 & 58.6 \\
            \midrule
            
            \multicolumn{11}{l}{\textit{Discriminative Models}} \\
            V-JEPA v2 \cite{assran2025v}  & 1B & 61.7 & 54.9  & 60.2 & 54.7 & 79.8  & 41.5  & 106.6 & 30.7 & 61.2 \\
            DinoV3-Large \cite{simeoni2025dinov3} & 0.3B & 61.1 & 54.2 & 59.6 & 54.1 & 80.6 & 41.1 & 105.9 & 30.5 & 61.9 \\
            VGGT \cite{wang2025vggt} & 1.1B & 62.3 & 55.3 & 60.1 & 54.5 & 82.8 & 42.0 & 105.8 & 30.5 & 61.4 \\
            \midrule
            
            \multicolumn{11}{l}{\textit{Generative Models}} \\
            Stable Video Diffusion \cite{blattmann2023stable} & 1.5B & 61.3 & 54.8 & 59.9 & 54.6 & 80.9  & 41.9 & 105.1 & 30.1 & 61.3 \\
            Stable Diffusion 2.1 \cite{rombach2022high} & 0.9B & 62.1 & 55.1  & 60.3 & 54.9 & 83.0 & 42.0 & 106.8 & 30.4 & 60.6 \\
            Vmem \cite{li2025vmem}      & 1.3B & 62.5 & 55.7 & 60.2 & 54.7 & 82.0 & 41.9 & 106.0 & 30.0 & 61.4 \\
            SEVA \cite{zhou2025stable} & 1.3B & 62.3 & 55.5 & 60.1 & 54.5 & 82.5 & 42.1 & 107.6 & 30.8 & 60.9 \\ 
            VAE \cite{rombach2022high}        & 0.08B & 62.0 & 55.1  & 60.3 & 54.8 & 83.7  & 42.3  & 106.0 & 30.5 & 61.4 \\
            Wan2.1-VACE \cite{jiang2025vace} & 1.3B & 62.2 & 55.3 & 60.3 & 55.0 & 82.8  & 42.7 & 107.8 & 31.0 & 61.8 \\
            Wan2.1-T2V \cite{wan2025wan} & 1.3B & 63.2 & 56.2 & 60.8 & 55.1 & 83.2 & 42.2 & 106.3 & 30.4 & 61.3 \\
            \bottomrule
        \end{tabular}
\vspace{-10pt}
\end{table}

\begin{wrapfigure}{r}{0.4\textwidth}
  \centering
  \vspace{-25pt}
  \includegraphics[width=\linewidth]{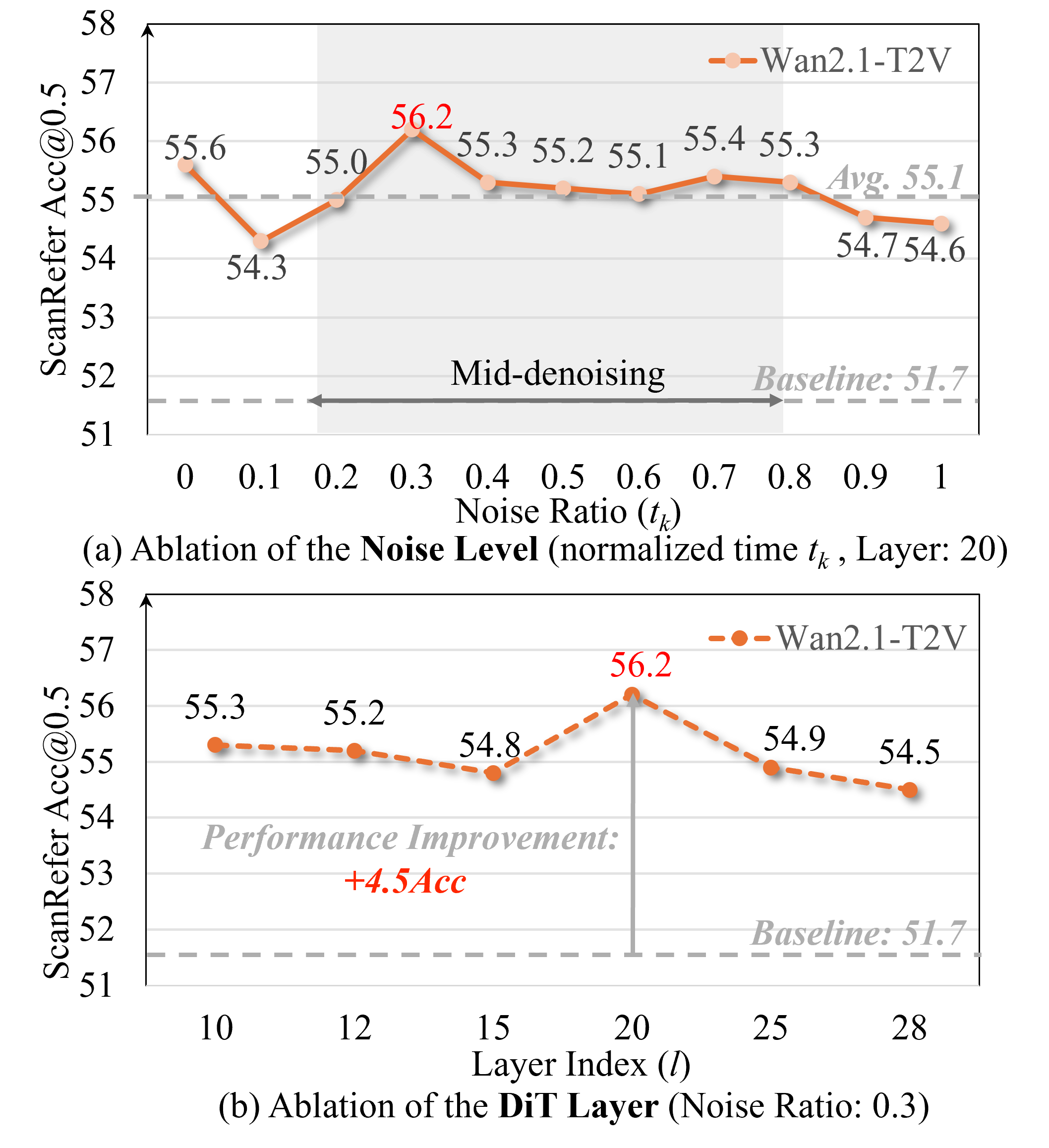} 
  \vspace{-15pt}
  \caption{\textbf{Ablation studies on noise injection and DiT depth. (a)} Performance peaks at intermediate noise levels. \textbf{(b)} Specific intermediate layers capture the most robust geometric cues.}
  \label{fig: dit_ablation}
  \vspace{-20pt}
\end{wrapfigure}

\myparagraph{Dynamics of Internal Representations: Noise Levels and Layers.}
\noindent The generative prior varies significantly across the diffusion process and network depth (Fig.~\ref{fig: dit_ablation}). \textbf{(i) Noise Ratio $t_k$:} Performance peaks at intermediate noise level. Clean latents underutilize the model's denoising capabilities, while excessive noise destroys structural signals. Moderate noise optimally forces the model to engage its learned physics to restore underlying 3D structures. \textbf{(ii) Layer Selection:} Intermediate layers provide the optimal abstraction for spatial reasoning, effectively balancing the low-level textures of early layers and the pixel-level rendering of deeper layers.

\noindent\textbf{Effectiveness of Adaptive Gated Fusion.}

Tab.~\ref{tab:feature_fusion} demonstrates the necessity of our Adaptive Gated Fusion. Relying solely on generative features causes a substantial performance drop, confirming that generative priors complement rather than replace semantic representations. Among lightweight fusion variants, our method achieves the best overall trade-off: it outperforms other fusion baselines on most metrics, and matches the best results on ScanQA (C/EM). We note that a naive Add operation attains a slightly higher SQA3D EM (61.8 vs.\ 61.3), but it is consistently weaker on grounding and captioning metrics, suggesting that a fixed, non-adaptive fusion weight cannot reliably resolve the semantic--generative distribution gap. In contrast, our token-level gating dynamically balances semantic and geometric priors, yielding more consistent gains across diverse tasks.

\begin{table*}[!t]
\centering
\scriptsize
\setlength{\tabcolsep}{0.8mm}
\caption{Ablation study of the effects of different feature fusion modules. All models are finetuned with the same training data and built on Wan2.1-T2V 1.3B at sampling step $k=300$ and the 20th layer feature.}
\vspace{-5pt}
\begin{tabular}{lccccccccc}
\toprule
\multirow{2}{*}{Type}
& \multicolumn{2}{c}{ScanRefer}
& \multicolumn{2}{c}{Multi3DRefer}
& \multicolumn{2}{c}{Scan2Cap}

& \multicolumn{2}{c}{ScanQA}
& SQA3D \\
\cmidrule(lr){2-3} \cmidrule(lr){4-5} \cmidrule(lr){6-7} \cmidrule(lr){8-9} \cmidrule(lr){10-10}
& Acc${.25}$ & Acc${.5}$ & F1${.25}$ & F1${.5}$ & C${.5}$ & B-4${.5}$ & C & EM & EM \\
\midrule
\rowcolor{baselinegray}
Baseline                       & 58.1 & 51.7 & 58.0 & 52.7 & \textbf{83.8} & 41.3 & 102.1 & 30.1 & 58.6 \\
\midrule
Only generative features       & 54.9 & 48.3 & 53.7 & 48.6 & 25.2 & 30.0 & 74.0 & 21.1 & 52.0 \\
\midrule
Add                            & 61.5 & 54.6 & 59.6 & 54.1 & 81.4 & 41.6 & \textbf{106.3} & \textbf{30.4} & \textbf{61.8} \\
Channel Concat+MLP                     & 55.1 & 48.9 & 53.6 & 48.7 & 33.2 & 31.8 & 81.6  & 22.9 & 52.3 \\
Sequence Concat                & 59.5 & 53.0 & 58.4 & 53.2 & 79.4 & 41.0 & 104.7 & 30.2 & 61.5 \\
Cross-Attn (1 Layer)           & 58.5 & 51.9 & 57.9 & 52.6 & 48.8 & 34.7 & 104.9 & 29.6 & 61.0 \\
Cross-Attn (3 Layers)          & 58.0 & 51.5 & 57.5 & 52.1 & 47.8 & 34.8 & 102.2 & 29.2 & 60.5 \\
Channel-Level-Gated            & 61.8 & 54.9 & 60.0 & 54.4 & 82.2 & 41.9 & 105.7 & 30.3 & 61.2 \\
Adaptive-Gated-Fusion(Ours)    & \textbf{63.2} & \textbf{56.2} & \textbf{60.8} & \textbf{55.1} & 83.2 & \textbf{42.2} & \textbf{106.3} & \textbf{30.4} & 61.3 \\
\bottomrule
\end{tabular}
\vspace{-10pt}
\label{tab:feature_fusion}
\end{table*}

\begin{figure*}[!t]
\centering
\includegraphics[width=0.99\textwidth]{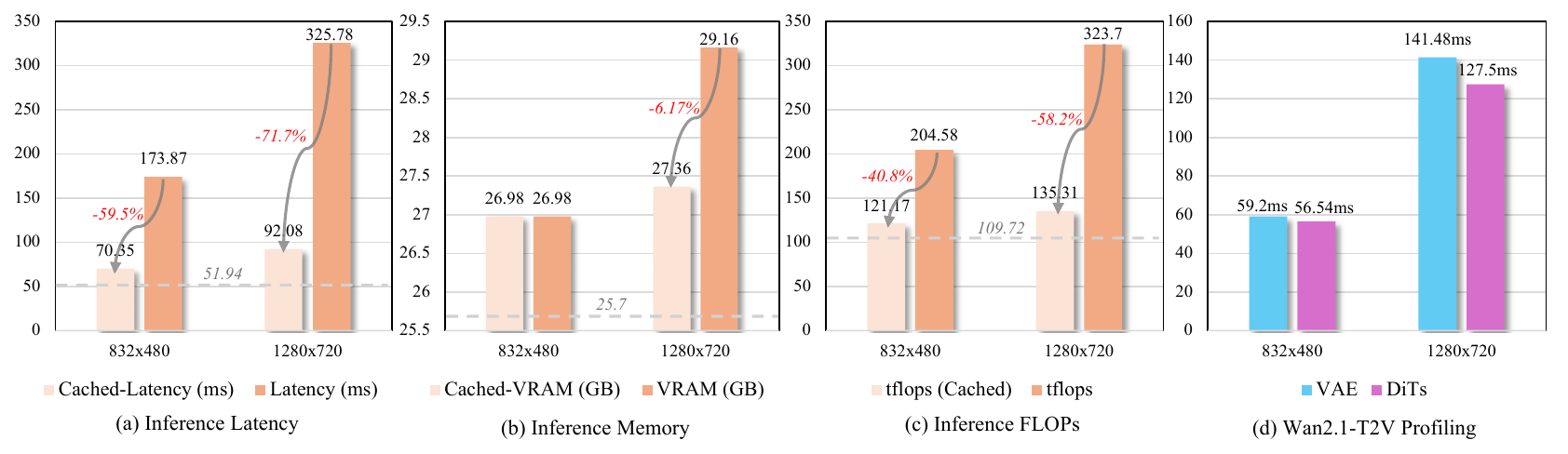}
\vspace{-10pt}
\caption{\textbf{Inference overhead} We cache the \textbf{Wan2.1-T2V} features once per scene and reuse them for all questions, substantially reducing inference overhead. The gray dashed line indicates the baseline result without the generative branch.}
\vspace{-15pt}
\label{fig: overhead}
\end{figure*}

\vspace{-10pt}

\section{Conclusion}
\label{sec:conclusion}

\vspace{-5pt}

We introduce \model{VEGA-3D}, a plug-and-play framework that repurposes modern video generation models as \textit{Latent World Simulators} to mitigate the spatial blindness of MLLMs.  By activating these priors via noise injection and aligning them with semantic tokens through \textit{Adaptive Gated Fusion}, VEGA-3D injects dense geometric anchors into MLLMs, consistently improving scene understanding, spatial reasoning, and manipulation \emph{without} extra 3D supervision.

\myparagraph{Limitations and Future Work.}
Incorporating a video diffusion backbone increases inference cost (Fig.~\ref{fig: overhead}). Nevertheless, it delivers substantial and consistent performance gains, making the trade-off acceptable in practice. Future work will distill these priors into lightweight encoders, extend the framework to more dynamic scene understanding, and scale the plug-and-play generative prior to larger Vision-Language-Action models for closed-loop embodied control.

\subsubsection*{Acknowledgements.}
This work was supported in part by the National Natural Science Foundation of China (NSFC) under Grants~62441615 and 623B2038, and in part by the Hubei Science and Technology Major Project under Grant~2024BAA007.

\clearpage
\appendix

\setcounter{figure}{0}
\renewcommand{\thefigure}{A\arabic{figure}}
\renewcommand{\theHfigure}{A\arabic{figure}}
\renewcommand{\theHtable}{A\arabic{table}}
\renewcommand{\theHsection}{appendix.\Alph{section}}
\renewcommand{\theHsubsection}{appendix.\Alph{section}.\arabic{subsection}}
\makeatletter
\renewcommand{\fnum@figure}{Fig-\thefigure}
\makeatother

\input{sec/appendix}

%
%
\bibliographystyle{splncs04}
\bibliography{main}

\end{document}

%% file: sec/appendix.tex
\setcounter{table}{0}
\renewcommand{\thetable}{A\arabic{table}}
\makeatletter
\renewcommand{\fnum@table}{Tab-\thetable}
\makeatother

\begin{center}
{\LARGE\bfseries Supplementary Material}
\end{center}
\vspace{-6pt}

\section{Technical Appendices}

This supplementary material provides additional technical details of our method, including the overall training procedure, the detailed configurations of the compared visual backbones, the implementation details of the multi-view correspondence score, and the normalized overall score (NOS) analysis.

Algorithm~\ref{alg:gen_fusion_train} summarizes the training pipeline of VEGA-3D. It highlights how the semantic branch and the generative branch are constructed, aligned, and fused before the resulting visual tokens are passed to the language model.

\begin{algorithm}
\scriptsize
\setlength{\abovecaptionskip}{3pt}
\setlength{\belowcaptionskip}{2pt}
\caption{Training Pipeline with a Generative Visual Encoder and Feature Fusion}
\label{alg:gen_fusion_train}
\begin{algorithmic}[1]
\Require Training set $\mathcal{D}=\{(x_i,q_i,a_i)\}_{i=1}^N$, sampled frame number $K$,
semantic encoder $E_{2D}$, semantic projector $P_{2D}$, LLM $M_\theta$,
generative source $\sigma \in \{\texttt{offline}, \texttt{online}\}$
\State Initialize $E_{2D}$, $P_{2D}$, and $M_\theta$ from the stage-1 checkpoint
\State Build generative projector $P_g$ and fusion module $F$
\If{$\sigma=\texttt{online}$}
    \State Keep the generative encoder $E_g$ frozen and instantiate it lazily at the first forward pass
\EndIf
\For{each minibatch $\mathcal{B}=\{(x_i,q_i,a_i)\}_{i=1}^B$}
    \For{each sample $(x_i,q_i,a_i)$ in $\mathcal{B}$}
        \State $\mathcal{V}_i \leftarrow \textsc{SampleFrames}(x_i, K)$
        \State $\mathcal{W}_i \leftarrow \textsc{UnprojectDepthToWorld}(\mathcal{V}_i)$
        \State $\mathbf{Z}^{2d}_i \leftarrow P_{2D}(E_{2D}(\mathcal{V}_i))$
        \State $\mathbf{Z}^{2d}_i \leftarrow \textsc{PoolTo14x14}(\mathbf{Z}^{2d}_i)$ \Comment{$196$ tokens per frame}
        \If{$\sigma=\texttt{offline}$}
            \State $\mathbf{G}_i \leftarrow \textsc{LoadOfflineGenerativeFeatures}(x_i)$
        \Else
            \State $\mathbf{G}_i \leftarrow \textsc{CacheOrEncode}(E_g, \mathcal{V}_i)$
            \Comment{inference mode, no gradient}
        \EndIf
        \State $\mathbf{Z}^{gen}_i \leftarrow P_g(\textsc{FlattenToTokens}(\mathbf{G}_i))$
        \State $\mathbf{c}_i \leftarrow \textsc{BuildInstructionContext}(q_i)$
        \Comment{optional, for instruction-aware gating}
        \State $\mathbf{Z}^{fuse}_i \leftarrow F(\mathbf{Z}^{2d}_i, \mathbf{Z}^{gen}_i, \mathbf{c}_i)$
        \State $\mathbf{Z}^{fuse}_i \leftarrow \mathbf{Z}^{fuse}_i + \textsc{3DPosEnc}(\mathcal{W}_i)$
        \State $\mathbf{X}^{vis}_i \leftarrow \textsc{SerializeGridTokens}(\mathbf{Z}^{fuse}_i)$
        \State $(\mathbf{e}_i,\mathbf{y}_i) \leftarrow \textsc{InsertVisualTokens}(q_i, a_i, \mathbf{X}^{vis}_i)$
        \Comment{replace the special image token with visual tokens}
    \EndFor
    \State $\mathcal{L}_{\mathrm{LM}} \leftarrow \frac{1}{B}\sum_{i=1}^{B}
    \textsc{CrossEntropy}(M_\theta(\mathbf{e}_i), \mathbf{y}_i)$
    \State $\mathcal{L} \leftarrow \mathcal{L}_{\mathrm{LM}}$
    \Comment{optionally plus grounding loss}
    \State Update the currently enabled trainable parameters $\Theta$ using $\nabla_{\Theta}\mathcal{L}$
\EndFor
\end{algorithmic}
\end{algorithm}

\section{More Implementation Details}

\subsection{Training Datasets}

We summarize the training data used in our three experimental settings: 3D scene understanding, spatial reasoning, and robotic manipulation.

\paragraph{3D scene understanding.}
For the 3D scene understanding setting, we follow the same training-data protocol as Video-3D LLM~\cite{video3dllm}. Specifically, the model is trained in a multi-task manner on the combination of five public benchmarks: ScanRefer~\cite{scanrefer}, Multi3DRefer~\cite{multi3drefer}, Scan2Cap~\cite{scan2cap}, ScanQA~\cite{scanqa}, and SQA3D~\cite{sqa3d}. Following Video-3D LLM, these tasks are all built from ScanNet scenes~\cite{scannet} and are converted into video-style multi-view inputs for unified training. This setting covers the three main categories studied in our paper, i.e., 3D visual grounding, dense captioning, and question answering. We do not introduce additional 3D instruction-tuning corpora beyond this mixed training set, so that the comparison with the Video-3D LLM baseline remains controlled and fair.

\paragraph{Spatial reasoning.}
For spatial reasoning, we adopt the \emph{S1} training set defined in VG-LLM~\cite{zheng2025learning}. In that work, S1 denotes the default mixed training data composed of sampled instances from SPAR-7M and the LLaVA-Hound split of LLaVA-Video-178K. SPAR-7M provides spatially enriched supervision spanning diverse 3D reasoning skills, while the LLaVA-Hound split is included to preserve the general video-language capability of the backbone. Following the VG-LLM setting, we use only this S1 mixture for training and do not incorporate the additional VLM-3R ~\cite{fan2025vlm} data used in their extended ablation study. Therefore, our spatial reasoning results isolate the gain brought by generative priors under the same core data setting, without relying on extra task-specific synthetic supervision.

\paragraph{Robotic manipulation.}
For robotic manipulation, we directly use the standard LIBERO benchmark~\cite{liu2024libero}. Following the protocol adopted by OpenVLA-OFT~\cite{kim2025fine}, we evaluate on the four canonical task suites: LIBERO-Spatial, LIBERO-Object, LIBERO-Goal, and LIBERO-Long. These suites are designed to test policy generalization under varying spatial layouts, object identities, goal conditions, and long-horizon task compositions, respectively. In our experiments, VEGA-3D is trained and evaluated on the same LIBERO downstream data as the OpenVLA-OFT baseline, without introducing additional robot datasets or auxiliary manipulation supervision, which keeps the comparison focused on the effect of our visual generative priors.

\subsection{Backbone Configurations for Tab.~4}

Tab.~\ref{tab:encoder_details} summarizes the detailed architecture and tokenization settings of the visual backbones compared in Tab.~4, including the discriminative models, the 3D foundation model, and the generative encoders. Unless otherwise noted, all extracted features are finally aligned to $14 \times 14 = 196$ spatial tokens before fusion with the semantic branch.

\begin{table*}
    \centering
    \setlength{\tabcolsep}{1pt}
    \caption{Detailed configurations of the visual backbones used in Tab.~4. ``Native Tok.'' denotes the token grid at the extracted feature stage before the final $14 \times 14$ alignment.}
    \label{tab:encoder_details}
    \resizebox{\linewidth}{!}{
    \begin{tabular}{l l l c c c c c}
        \toprule
        Model & Group & Architecture & Input Res. & Feature Stage & Patch Size & Native Tok. & Hidden Size \\
        \midrule
        \multicolumn{8}{l}{\textit{Discriminative / self-supervised models}} \\
        DINOv3-Large \cite{simeoni2025dinov3} & Self-supervised & ViT-L/16 & $224 \times 224$ & final patch tokens & 16 & 196 & 1024 \\
        V-JEPA v2 \cite{assran2025v} & Self-supervised & ViT-G/16 & $224 \times 224$ & encoder patch tokens & 16 & 196 & 1408 \\
        \midrule
        \multicolumn{8}{l}{\textit{3D foundation model}} \\
        VGGT \cite{wang2025vggt} & Geometry & ViT-L/14 + aggregator & $196 \times 196$ & last aggregator patch tokens & 14 & 196 & 2048 \\
        \midrule
        \multicolumn{8}{l}{\textit{Generative models}} \\
        Stable Video Diffusion \cite{blattmann2023stable} & Video diffusion & VAE + UNet & $896 \times 896$ & UNet down block output & 64 & 196 & 1280 \\
        Stable Diffusion 2.1 \cite{rombach2022high} & Image diffusion & VAE + UNet & $896 \times 896$ & UNet pre-midblock feature & 64 & 196 & 1280 \\
        Vmem \cite{li2025vmem} & Video diffusion & VAE + UNet & $896 \times 896$ & final input-block feature & 64 & 196 & 1280 \\
        SEVA \cite{zhou2025stable} & Video diffusion & VAE + UNet & $896 \times 896$ & final input-block feature & 64 & 196 & 1280 \\
        VAE \cite{rombach2022high} & Autoencoder & VAE & $224 \times 224$ & latent feature & 16 & 196 & 4 \\
        Wan2.1-VACE \cite{jiang2025vace} & Video diffusion & VAE + DiT & $1280 \times 720$ & 20th DiT block feature & 16 & 3600 & 1536 \\
        Wan2.1-T2V \cite{wan2025wan} & Video diffusion & VAE + DiT & $1280 \times 720$ & 20th DiT block feature & 16 & 3600 & 1536 \\
        \bottomrule
    \end{tabular}
    }
\end{table*}

\section{Details of  Multi-view Correspondence Score and Normalized Overall Score}

\subsection{Multi-view Correspondence Score}

The multi-view correspondence score used in Fig.~3 is not computed by exhaustively matching arbitrary token pairs from different frames. Instead, the implementation first associates each visual token with a 3D point and then only compares tokens that are assigned to the same voxel and come from different views.

Concretely, for each scene we uniformly sample up to 32 frames. Depth maps are unprojected into 3D world coordinates using the camera intrinsics and poses, and the poses are first transformed by the scene-specific axis-alignment matrix so that all views are expressed in a shared scene coordinate system. In the default analysis setting, the sampled images are resized to $384 \times 384$ before feature extraction, and the dense world coordinates are resized with the same image transform. After extracting a feature map $\mathbf{F} \in \mathbb{R}^{T \times C \times H_f \times W_f}$, the world coordinates are temporally resampled if needed and then adaptively average-pooled to the same $H_f \times W_f$ grid; in the default encoder setting, this corresponds to a $14 \times 14$ token grid. Therefore, each feature token is paired with one pooled 3D point rather than with all raw pixels inside the corresponding image patch.

Let $f_n \in \mathbb{R}^C$ denote the flattened token feature, $x_n \in \mathbb{R}^3$ denote its pooled 3D coordinate, and $v(n)$ denote its view index. The scene-specific voxel index is defined as
\begin{equation}
g(n)=\left\lfloor \frac{x_n - x_{\min}}{s} \right\rfloor,
\end{equation}
where $x_{\min}$ is the element-wise minimum of all token coordinates in the current scene and $s$ is the voxel size, which is set to $0.1$ m by default. For each voxel $k$ and view $t$, we collect
\begin{equation}
\Omega_{k,t}=\{n \mid g(n)=k,\ v(n)=t\}.
\end{equation}
If $\Omega_{k,t}$ is non-empty, all tokens from the same view inside that voxel are first averaged and then L2-normalized to form a per-view prototype:
\begin{equation}
p_{k,t}=\mathrm{Normalize}\left(\frac{1}{|\Omega_{k,t}|}\sum_{n\in\Omega_{k,t}} f_n\right).
\end{equation}
Only voxels observed by at least two distinct views contribute to the final score. Denoting the set of such views for voxel $k$ by $V_k$, the cross-view correspondence terms are
\begin{equation}
c_{k,t,t'} = p_{k,t}^{\top} p_{k,t'}, \qquad t<t',\ t,t' \in V_k.
\end{equation}
The scene-level multi-view correspondence score is then computed as
\begin{equation}
S = \frac{\sum_k \sum_{t<t',\ t,t'\in V_k} c_{k,t,t'}}{\sum_k \binom{|V_k|}{2}}.
\end{equation}

Several implementation details are important for reproducing the exact score. First, this is a \emph{pair-weighted} average rather than a voxel-weighted average: a voxel observed by more views contributes more cross-view pairs. Second, tokens from the same view are never self-matched; multiple same-view tokens that fall into the same voxel are merged into one prototype before normalization. Third, the main correspondence script does not explicitly discard invalid depth pixels with zero depth, which differs from the visualization code path that applies an additional validity mask. Fourth, if a scene contains no voxel observed by at least two views, its score is recorded as NaN. Finally, dataset-level statistics are computed by first obtaining one score per scene and then reporting the mean and standard deviation over valid scene scores, rather than by globally aggregating all token pairs from all scenes into a single pool.

\subsection{Normalized Overall Score (NOS)}

To summarize the overall downstream performance with a single scalar, we compute a Normalized Overall Score (NOS) by first normalizing each metric to $[0,1]$ and then averaging across all metrics. Importantly, the discriminative models and the generative models are normalized \emph{separately}. In each group, we include the corresponding baseline when computing the per-metric minimum and maximum.

For a model $t$ and metric $m$, the normalized score is defined as 
\begin{equation}
\text{Norm}(x_{t,m}) =
\frac{x_{t,m} - \min\limits_{t' \in \mathcal{G}} x_{t',m}}
{\max\limits_{t' \in \mathcal{G}} x_{t',m} - \min\limits_{t' \in \mathcal{G}} x_{t',m}},
\end{equation}
where $\mathcal{G}$ denotes either the discriminative-model group or the generative-model group together with the baseline. We then compute
\begin{equation}
\text{NOS}(t) = \frac{100}{M}\sum_{m=1}^{M}\text{Norm}(x_{t,m}),
\end{equation}
where the final factor of 100 converts the averaged normalized score into a percentage. Here $M{=}9$ is the number of evaluation metrics used in Tab.~4, namely the ScanRefer accuracies at IoU thresholds 0.25 and 0.5 (Acc@0.25 and Acc@0.5), the Multi3DRefer F1 scores at IoU thresholds 0.25 and 0.5 (F1@0.25 and F1@0.5), the Scan2Cap captioning scores measured by CIDEr@0.5 and BLEU-4@0.5, the ScanQA question answering scores measured by CIDEr and exact match (EM), and the SQA3D exact match score.

Tab.~\ref{tab:nos_models} summarizes the final NOS values together with the multi-view correspondence scores used in the analysis. Since the raw task metrics are already reported in Tab.~4 of the main paper, we omit those dataset-wise results here.

\begin{table*}
    \centering
    \scriptsize
    \setlength{\tabcolsep}{1.6mm}
    \caption{Summary of NOS and multi-view correspondence scores for the models in Tab.~4. Both NOS and correspondence scores are reported in percentage. NOS is computed separately for the discriminative group and the generative group, each together with the baseline; therefore, the baseline row reports two NOS values, in the order of discriminative / generative normalization. Bold denotes the best result in each group.}
    \vspace{-5pt}
    \label{tab:nos_models}
    \begin{tabular}{lcc}
        \toprule
        Models & NOS Score (\%) & Correspondence Score (\%) \\
        \midrule

        \rowcolor{baselinegray}
        Baseline & 13.58 / 12.22 & - \\
        \midrule

        \multicolumn{3}{l}{\textit{Discriminative Models}} \\
        V-JEPA v2 \cite{assran2025v} & 77.54 & 72.00 \\
        DINOv3-Large \cite{simeoni2025dinov3} & 61.63 & 61.90 \\
        VGGT \cite{wang2025vggt} & \textbf{88.24} & \textbf{77.21} \\
        \midrule

        \multicolumn{3}{l}{\textit{Generative Models}} \\
        Stable Video Diffusion \cite{blattmann2023stable} & 52.06 & 17.95 \\
        Stable Diffusion 2.1 \cite{rombach2022high} & 70.57 & 23.83 \\
        Vmem \cite{li2025vmem} & 63.75 & 66.74 \\
        SEVA \cite{zhou2025stable} & 75.28 & 76.15 \\
        VAE \cite{rombach2022high} & 77.29 & 79.69 \\
        Wan2.1-VACE \cite{jiang2025vace} & \textbf{89.32} & \textbf{97.04} \\
        Wan2.1-T2V \cite{wan2025wan} & 82.41 & 96.88 \\
        \bottomrule
    \end{tabular}
\end{table*}

\section{Probe analysis of the 3D priors in Generation Models}

\noindent Fig.~6 studies how the extracted generative prior changes with the diffusion timestep and the DiT depth. For completeness, we report the exact downstream results corresponding to Fig.~6 in Tabs.~\ref{tab:dit_timestep_ablation} and \ref{tab:dit_block_ablation}. We also provide additional timestep ablations for SEVA and Vmem in Tab.~\ref{tab:unet_timestep_ablation}, which show a similar trend: intermediate timesteps generally yield stronger and more stable downstream performance than very late timesteps.

\begin{table*}[htbp]
    \centering
    \scriptsize
    \setlength{\tabcolsep}{0.8mm}
    \caption{Exact ablation results for different diffusion timesteps in Fig.~6(a). All rows use Wan2.1-T2V-1.3B, and only the sampling timestep is varied.}
    \vspace{-5pt}
    \label{tab:dit_timestep_ablation}
    \resizebox{\linewidth}{!}{
    \begin{tabular}{lccccccccc}
        \toprule
        \multirow{2}{*}{Timestep}
        & \multicolumn{2}{c}{ScanRefer}
        & \multicolumn{2}{c}{Multi3DRefer}
        & \multicolumn{2}{c}{Scan2Cap}
        & \multicolumn{2}{c}{ScanQA}
        & SQA3D \\
        \cmidrule(lr){2-3} \cmidrule(lr){4-5} \cmidrule(lr){6-7} \cmidrule(lr){8-9} \cmidrule(lr){10-10}
        & Acc$_{.25}$ & Acc$_{.5}$ & F1$_{.25}$ & F1$_{.5}$ & C$_{.5}$ & B-4$_{.5}$ & C & EM & EM \\
        \midrule
        \rowcolor{baselinegray}
        Baseline & 58.1 & 51.7 & 58.0 & 52.7 & 83.8 & 41.3 & 102.1 & 30.1 & 58.6 \\
        0 & 62.4 & 55.6 & 60.6 & 55.0 & 82.3 & 42.0 & 106.3 & 30.2 & 60.9 \\
        100 & 61.3 & 54.3 & 60.0 & 54.4 & 81.7 & 41.8 & 106.1 & 30.0 & 61.5 \\
        200 & 61.9 & 55.0 & 60.1 & 54.7 & 83.1 & 42.4 & 107.6 & 30.6 & 60.7 \\
        300 & 63.2 & 56.2 & 60.8 & 55.1 & 83.2 & 42.2 & 106.3 & 30.4 & 61.3 \\
        400 & 62.4 & 55.3 & 60.3 & 54.6 & 84.3 & 42.6 & 105.1 & 29.9 & 60.7 \\
        500 & 61.8 & 55.2 & 59.8 & 54.4 & 81.7 & 42.1 & 106.6 & 31.1 & 61.4 \\
        600 & 62.1 & 55.1 & 60.2 & 54.6 & 83.2 & 42.2 & 105.0 & 30.1 & 61.1 \\
        700 & 62.2 & 55.4 & 60.5 & 55.0 & 80.6 & 41.7 & 105.6 & 30.3 & 61.2 \\
        800 & 62.1 & 55.3 & 60.0 & 54.6 & 82.7 & 41.7 & 106.4 & 30.5 & 60.7 \\
        900 & 61.6 & 54.7 & 59.7 & 54.3 & 81.4 & 41.9 & 107.7 & 30.8 & 61.0 \\
        1000 & 61.4 & 54.6 & 59.7 & 54.3 & 82.0 & 42.1 & 105.0 & 30.2 & 60.9 \\
        \bottomrule
    \end{tabular}
    }
\end{table*}

\begin{table*}[htbp]
    \centering
    \scriptsize
    \setlength{\tabcolsep}{0.8mm}
    \caption{Additional timestep ablations for SEVA ~\cite{zhou2025stable} and Vmem ~\cite{li2025vmem}. Rows are grouped by backbone, and only the sampling timestep varies within each group. Similar to Wan2.1-T2V, both backbones tend to perform better at intermediate timesteps than at the end of the diffusion trajectory.}
    \vspace{-5pt}
    \label{tab:unet_timestep_ablation}
    \resizebox{\linewidth}{!}{
    \begin{tabular}{lccccccccc}
        \toprule
        \multirow{2}{*}{Timestep}
        & \multicolumn{2}{c}{ScanRefer}
        & \multicolumn{2}{c}{Multi3DRefer}
        & \multicolumn{2}{c}{Scan2Cap}
        & \multicolumn{2}{c}{ScanQA}
        & SQA3D \\
        \cmidrule(lr){2-3} \cmidrule(lr){4-5} \cmidrule(lr){6-7} \cmidrule(lr){8-9} \cmidrule(lr){10-10}
        & Acc$_{.25}$ & Acc$_{.5}$ & F1$_{.25}$ & F1$_{.5}$ & C$_{.5}$ & B-4$_{.5}$ & C & EM & EM \\
        \midrule
        \multicolumn{10}{l}{\textit{SEVA}} \\
        250 & 61.3 & 54.5 & 59.9 & 54.5 & 81.6 & 41.7 & 105.3 & 30.3 & 62.0 \\
        300 & 60.9 & 54.2 & 59.5 & 54.2 & 79.5 & 41.1 & 106.3 & 30.4 & 60.7 \\
        400 & 62.3 & 55.4 & 60.4 & 54.9 & 80.4 & 41.4 & 105.4 & 30.2 & 60.8 \\
        500 & 62.1 & 55.1 & 60.3 & 54.7 & 83.1 & 42.3 & 105.7 & 30.3 & 61.3 \\
        600 & 61.6 & 54.7 & 60.4 & 54.8 & 81.8 & 41.9 & 105.3 & 30.2 & 61.8 \\
        700 & 62.3 & 55.5 & 60.1 & 54.5 & 82.5 & 42.1 & 107.6 & 30.8 & 60.9 \\
        800 & 61.3 & 54.6 & 59.8 & 54.4 & 80.9 & 41.8 & 105.5 & 30.6 & 61.8 \\
        900 & 61.8 & 55.0 & 60.1 & 54.6 & 83.4 & 42.4 & 107.2 & 30.7 & 61.2 \\
        1000 & 61.4 & 54.3 & 59.6 & 54.1 & 81.6 & 42.0 & 105.7 & 30.7 & 61.1 \\
        \midrule
        \multicolumn{10}{l}{\textit{Vmem}} \\
        250 & 62.5 & 55.7 & 60.5 & 54.9 & 80.9 & 41.5 & 105.8 & 30.3 & 61.7 \\
        300 & 62.5 & 55.7 & 60.2 & 54.7 & 82.0 & 41.9 & 106.0 & 29.9 & 61.4 \\
        400 & 62.0 & 55.0 & 60.4 & 54.9 & 81.1 & 41.7 & 105.6 & 30.3 & 60.7 \\
        500 & 62.1 & 55.2 & 60.2 & 54.6 & 81.8 & 41.6 & 104.8 & 30.0 & 60.8 \\
        600 & 62.3 & 55.5 & 60.5 & 55.0 & 81.4 & 41.7 & 105.5 & 30.4 & 61.4 \\
        700 & 62.0 & 55.0 & 60.2 & 54.6 & 81.3 & 41.8 & 105.3 & 30.1 & 61.4 \\
        800 & 61.2 & 54.3 & 59.3 & 53.7 & 80.6 & 41.4 & 107.3 & 30.7 & 60.7 \\
        900 & 61.9 & 54.9 & 60.1 & 54.6 & 81.3 & 42.1 & 107.1 & 30.3 & 61.5 \\
        1000 & 61.7 & 54.8 & 59.8 & 54.2 & 82.3 & 41.7 & 107.3 & 30.5 & 61.7 \\
        \bottomrule
    \end{tabular}
    }
\end{table*}

\begin{table*}[htbp]
    \centering
    \scriptsize
    \setlength{\tabcolsep}{0.8mm}
    \caption{Exact ablation results for different DiT blocks in Fig.~6(b). All rows use Wan2.1-T2V-1.3B at timestep $k=300$ under the $1280 \times 720$ setting, and only the extracted DiT block is varied.}
    \vspace{-5pt}
    \label{tab:dit_block_ablation}
    \resizebox{\linewidth}{!}{
    \begin{tabular}{lccccccccc}
        \toprule
        \multirow{2}{*}{Block}
        & \multicolumn{2}{c}{ScanRefer}
        & \multicolumn{2}{c}{Multi3DRefer}
        & \multicolumn{2}{c}{Scan2Cap}
        & \multicolumn{2}{c}{ScanQA}
        & SQA3D \\
        \cmidrule(lr){2-3} \cmidrule(lr){4-5} \cmidrule(lr){6-7} \cmidrule(lr){8-9} \cmidrule(lr){10-10}
        & Acc$_{.25}$ & Acc$_{.5}$ & F1$_{.25}$ & F1$_{.5}$ & C$_{.5}$ & B-4$_{.5}$ & C & EM & EM \\
        \midrule
        10 & 62.3 & 55.3 & 60.4 & 54.8 & 83.4 & 42.6 & 106.8 & 30.7 & 61.1 \\
        12 & 61.9 & 55.2 & 60.6 & 54.9 & 81.2 & 41.3 & 106.0 & 30.3 & 61.7 \\
        15 & 61.8 & 54.8 & 60.0 & 54.4 & 81.6 & 41.4 & 105.8 & 30.4 & 60.9 \\
        20 & 63.2 & 56.2 & 60.8 & 55.1 & 83.2 & 42.2 & 106.3 & 30.4 & 61.3 \\
        25 & 61.9 & 54.9 & 60.0 & 54.5 & 81.2 & 41.5 & 105.5 & 30.1 & 61.3 \\
        28 & 61.4 & 54.5 & 59.4 & 53.9 & 84.0 & 42.4 & 106.0 & 30.0 & 61.6 \\
        \bottomrule
    \end{tabular}
    }
\end{table*}

\section{Additional Qualitative Results}

\begin{figure*}[htbp]
\centering
\includegraphics[width=\textwidth]{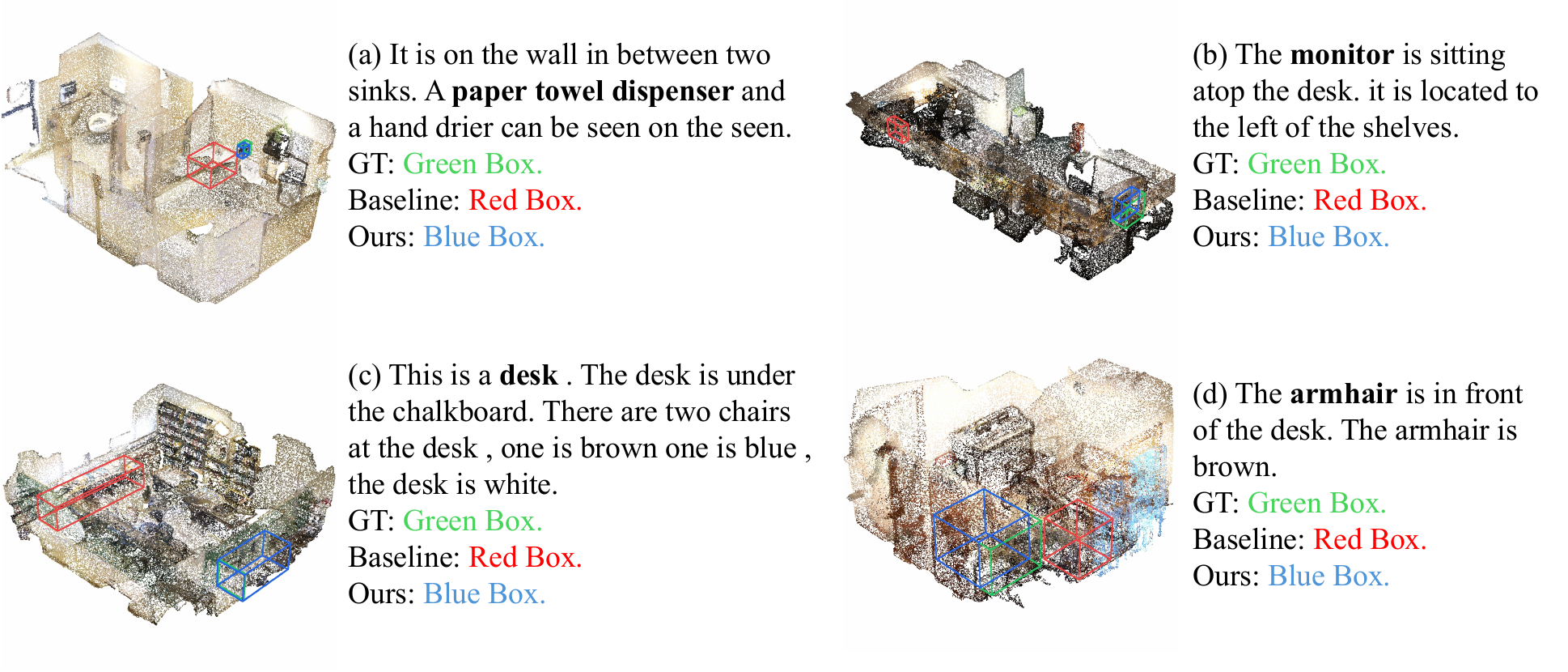}
\caption{Qualitative comparison on ScanRefer. Compared with the baseline, VEGA-3D localizes the referred object accurately under clutter, occlusion, and ambiguous referring expressions, reflecting stronger spatial grounding from the generative prior.}
\label{fig:scanrefer_cases}
\end{figure*}

\begin{figure*}[htbp]
\centering
\includegraphics[width=\textwidth]{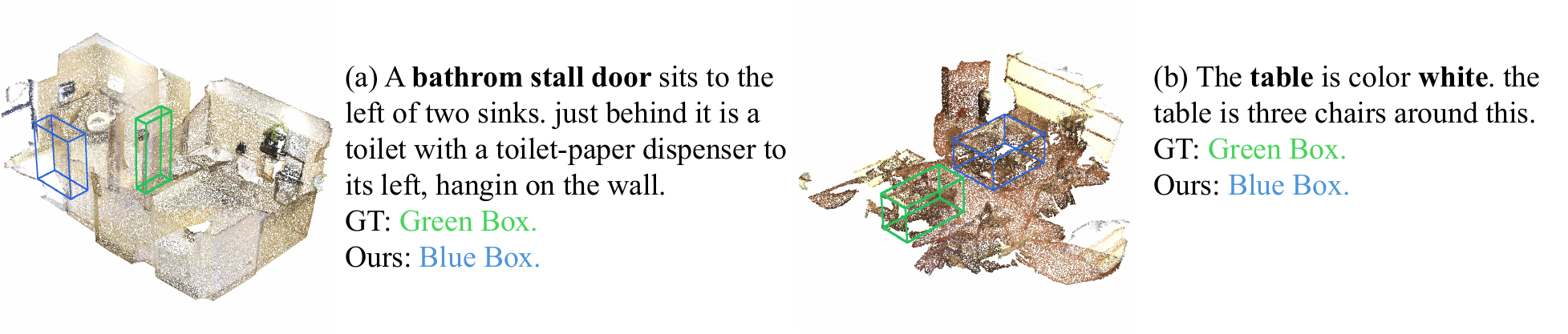}
\caption{Representative failure case on ScanRefer. We show the VEGA-3D prediction and the ground-truth box, indicating that VEGA-3D captures a reasonable spatial anchor but can still struggle with fine-grained instance disambiguation in cluttered scenes.}
\label{fig:scanrefer_failure_cases}
\end{figure*}

\begin{figure*}[htbp]
\centering
\includegraphics[width=\textwidth]{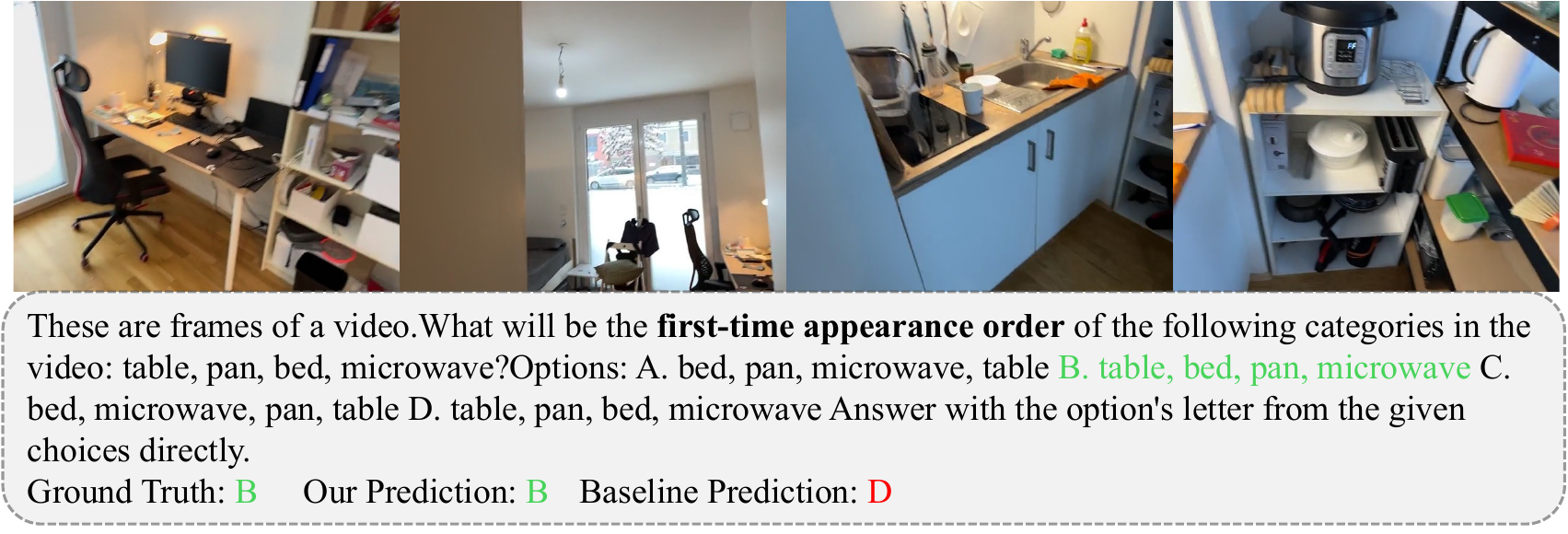}
\caption{Qualitative comparison on the Appearance Order subset of VSI-Bench. VEGA-3D better captures the ordering of object appearances and is less distracted by locally plausible but temporally inconsistent choices than the baseline.}
\label{fig:vsi_appearance_order}
\end{figure*}

\begin{figure*}[htbp]
\centering
\includegraphics[width=\textwidth]{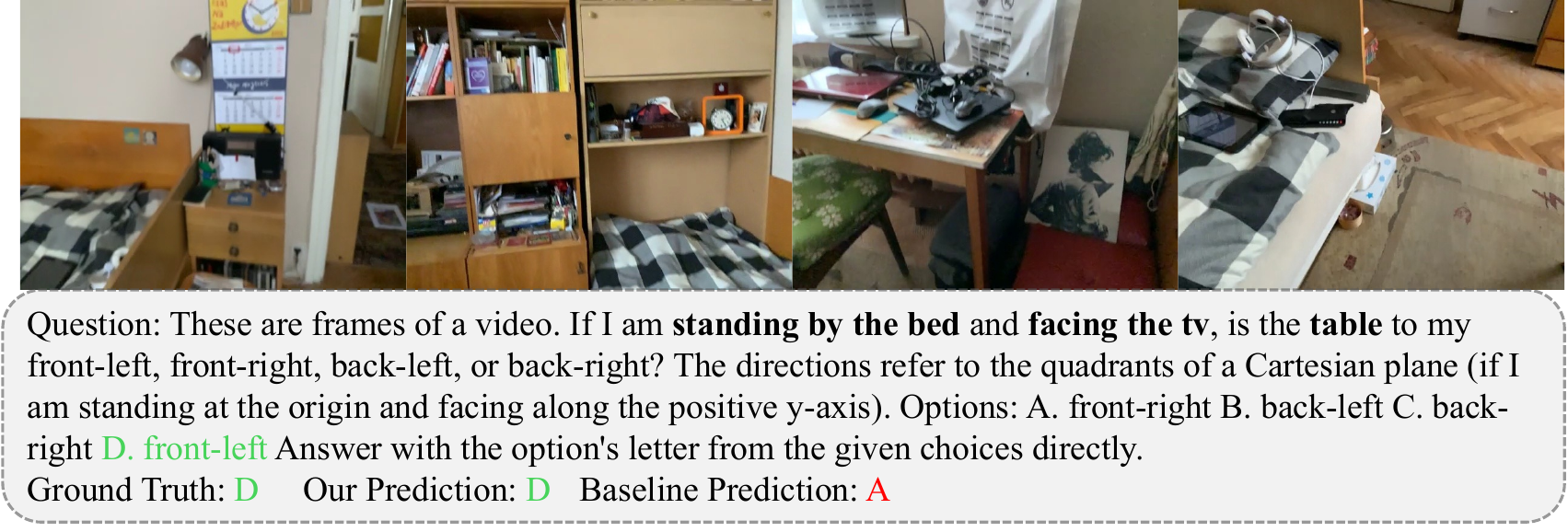}
\caption{Qualitative comparison on the Relative Direction subset of VSI-Bench. VEGA-3D yields reliable directional reasoning, such as left/right and front/behind relations, by grounding the decision in geometry-consistent visual features.}
\label{fig:vsi_rel_direction}
\end{figure*}

\begin{figure*}[htbp]
\centering
\includegraphics[width=\textwidth]{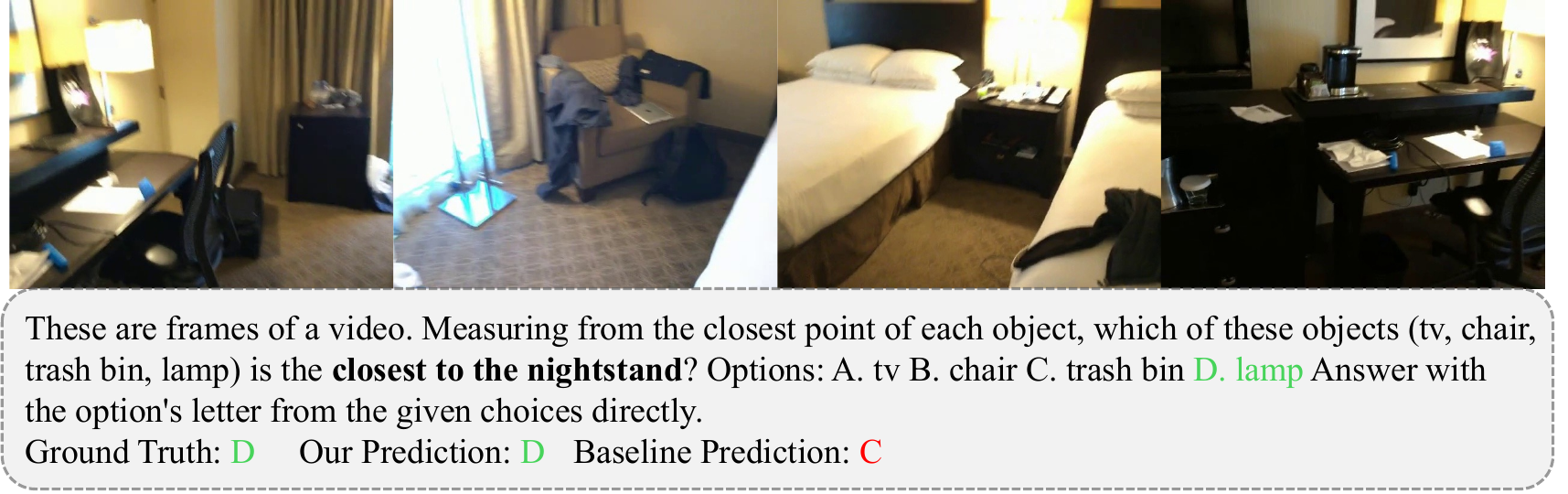}
\caption{Qualitative comparison on the Relative Distance subset of VSI-Bench. VEGA-3D distinguishes near/far relationships and relative depth ordering, whereas the baseline is prone to confuse objects with similar semantics but different spatial positions.}
\label{fig:vsi_rel_distance}
\end{figure*}

We provide additional qualitative comparisons between VEGA-3D and the corresponding baseline models on ScanRefer and representative VSI-Bench sub-tasks. The success cases in Figs.~\ref{fig:scanrefer_cases}, \ref{fig:vsi_appearance_order}, \ref{fig:vsi_rel_direction}, and \ref{fig:vsi_rel_distance} show that our method produces more spatially grounded predictions: on ScanRefer, it localizes the target object more precisely in cluttered indoor scenes, and on VSI-Bench it handles appearance order, relative direction, and relative distance more reliably. We also include a representative ScanRefer failure case in Fig.~\ref{fig:scanrefer_failure_cases}. Even when VEGA-3D does not identify the exact target instance, its prediction remains close to the ground-truth region, suggesting that the generative prior substantially improves coarse spatial grounding while fine-grained disambiguation among nearby similar objects is still challenging.

\section{Limitations}

Our study also highlights several limitations. First, generative priors are complementary to semantic features rather than a replacement for them: they help most on localization-centric and geometry-sensitive tasks, while semantic-heavy metrics such as captioning may improve less consistently. Second, not all generative backbones are equally suitable. In our analysis, DiT-based models exhibit substantially stronger multi-view consistency than several UNet-based alternatives, suggesting that transfer quality depends on the backbone architecture and pretraining regime. Third, incorporating a frozen video generator increases memory and inference cost, even though feature caching alleviates the overhead in practice. Finally, the best extraction setting currently depends on manually chosen intermediate timesteps and feature layers, and our experiments are still centered on indoor multi-view settings. Future work will study how to distill these priors into lighter encoders, learn more adaptive extraction strategies, and extend the framework to more dynamic and open-world environments.

%% file: main.bib
@string{CVPR = "Proc. IEEE Conf. Comput. Vis. Pattern Recognit."}

@string{ICCV = "Proc. IEEE Int. Conf. Comput. Vis."}

@string{ECCV = "Proc. Eur. Conf. Comput. Vis."}

@string{WACV = "Proc. IEEE Winter Conf. Appl. Comput. Vis."}

@string{NIPS = "Proc. Adv. Neural Inf. Process. Syst."}

@string{NeurIPS = "Proc. Adv. Neural Inf. Process. Syst."}

@string{ICML = "Proc. Int. Conf. Mach. Learn."}

@string{ICLR = "Proc. Int. Conf. Learn. Representations"}

@string{ICRA   = "Proc. IEEE Int. Conf. Robotics Automation"}

@string{ACL = "Proc. Annual Meeting of the Association for Computational Linguistics"}

@inproceedings{vg-llm,
  title={Learning from Videos for 3D World: Enhancing MLLMs with 3D Vision Geometry Priors},
  author={Zheng, Duo and Huang, Shijia and Li, Yanyang and Wang, Liwei},
  booktitle=NIPS,
  year={2025}
}

@article{Point-bind,
  title={Point-bind \& point-llm: Aligning point cloud with multi-modality for 3d understanding, generation, and instruction following},
  author={Guo, Ziyu and Zhang, Renrui and Zhu, Xiangyang and Tang, Yiwen and Ma, Xianzheng and Han, Jiaming and Chen, Kexin and Gao, Peng and Li, Xianzhi and Li, Hongsheng and others},
  journal={arXiv preprint arXiv:2309.00615},
  year={2023}
}

@inproceedings{inst3d,
  title={Inst3d-lmm: Instance-aware 3d scene understanding with multi-modal instruction tuning},
  author={Yu, Hanxun and Li, Wentong and Wang, Song and Chen, Junbo and Zhu, Jianke},
  booktitle=CVPR,
  year={2025}
}

@inproceedings{gpt4point,
  title={Gpt4point: A unified framework for point-language understanding and generation},
  author={Qi, Zhangyang and Fang, Ye and Sun, Zeyi and Wu, Xiaoyang and Wu, Tong and Wang, Jiaqi and Lin, Dahua and Zhao, Hengshuang},
  booktitle=CVPR,
  year={2024}
}

@article{chat3d,
  title={Chat-3d: Data-efficiently tuning large language model for universal dialogue of 3d scenes},
  author={Wang, Zehan and Huang, Haifeng and Zhao, Yang and Zhang, Ziang and Zhao, Zhou},
  journal={arXiv preprint arXiv:2308.08769},
  year={2023}
}

@inproceedings{3dllava,
  title={3d-llava: Towards generalist 3d lmms with omni superpoint transformer},
  author={Deng, Jiajun and He, Tianyu and Jiang, Li and Wang, Tianyu and Dayoub, Feras and Reid, Ian},
  booktitle=CVPR,
  year={2025}
}

@inproceedings{video3dllm,
  title={Video-3d llm: Learning position-aware video representation for 3d scene understanding},
  author={Zheng, Duo and Huang, Shijia and Wang, Liwei},
  booktitle=CVPR,
  year={2025}
}

@article{siglip2,
  title={Siglip 2: Multilingual vision-language encoders with improved semantic understanding, localization, and dense features},
  author={Tschannen, Michael and Gritsenko, Alexey and Wang, Xiao and Naeem, Muhammad Ferjad and Alabdulmohsin, Ibrahim and Parthasarathy, Nikhil and Evans, Talfan and Beyer, Lucas and Xia, Ye and Mustafa, Basil and others},
  journal={arXiv preprint arXiv:2502.14786},
  year={2025}
}

@inproceedings{mvt,
 author = {Shijia Huang and
Yilun Chen and
Jiaya Jia and
Liwei Wang},
 booktitle = CVPR,
 title = {Multi-View Transformer for 3D Visual Grounding},
 year = {2022}
}

@inproceedings{scanrefer,
  title={Scanrefer: 3d object localization in rgb-d scans using natural language},
  author={Chen, Dave Zhenyu and Chang, Angel X and Nie{\ss}ner, Matthias},
  booktitle=ECCV,
  year={2020},
  note = {License: Creative Commons Attribution-NonCommercial-ShareAlike 3.0 Unported License}
}

@inproceedings{ll3da,
 author = {Sijin Chen and
Xin Chen and
Chi Zhang and
Mingsheng Li and
Gang Yu and
Hao Fei and
Hongyuan Zhu and
Jiayuan Fan and
Tao Chen},
 booktitle = CVPR,
 title = {{LL3DA:} Visual Interactive Instruction Tuning for Omni-3D Understanding,
Reasoning, and Planning},
 year = {2024}
}

@inproceedings{gpt4scene,
  title={Gpt4scene: Understand 3d scenes from videos with vision-language models},
  author={Qi, Zhangyang and Zhang, Zhixiong and Fang, Ye and Wang, Jiaqi and Zhao, Hengshuang},
  booktitle = ICLR,
  year={2026}
}

@article{grounded-3dllm,
 title={Grounded 3d-llm with referent tokens},
  author={Chen, Yilun and Yang, Shuai and Huang, Haifeng and Wang, Tai and Xu, Runsen and Lyu, Ruiyuan and Lin, Dahua and Pang, Jiangmiao},
  journal={arXiv preprint arXiv:2405.10370},
  year={2024}
}

@inproceedings{multi3drefer,
 author = {Yiming Zhang and
ZeMing Gong and
Angel X. Chang},
 booktitle = ICCV,
 title = {Multi3DRefer: Grounding Text Description to Multiple 3D Objects},
 year = {2023},
note    = {License: MIT}
}

@inproceedings{scan2cap,
  title={Scan2cap: Context-aware dense captioning in rgb-d scans},
  author={Chen, Zhenyu and Gholami, Ali and Nie{\ss}ner, Matthias and Chang, Angel X},
  booktitle=CVPR,
  year={2021},
note={License: Creative Commons Attribution-NonCommercial-ShareAlike 3.0 Unported License}
}

@inproceedings{scanqa,
 author = {Daichi Azuma and
Taiki Miyanishi and
Shuhei Kurita and
Motoaki Kawanabe},
 booktitle = CVPR,
 title = {ScanQA: 3D Question Answering for Spatial Scene Understanding},
 year = {2022},
note={License: Creative Commons Attribution-NonCommercial-ShareAlike 3.0 Unported License}
}

@inproceedings{3dvista,
 author = {Ziyu Zhu and
Xiaojian Ma and
Yixin Chen and
Zhidong Deng and
Siyuan Huang and
Qing Li},
 booktitle = ICCV,
 title = {3D-VisTA: Pre-trained Transformer for 3D Vision and Text Alignment},
 year = {2023}
}

@inproceedings{vil3drel,
 author = {Shizhe Chen and
Pierre{-}Louis Guhur and
Makarand Tapaswi and
Cordelia Schmid and
Ivan Laptev},
  booktitle=NIPS,
 title = {Language Conditioned Spatial Relation Reasoning for 3D Object Grounding},
 year = {2022}
}

@inproceedings{3dvg-trans,
 author = {Lichen Zhao and
Daigang Cai and
Lu Sheng and
Dong Xu},
 booktitle = ICCV,
 title = {3DVG-Transformer: Relation Modeling for Visual Grounding on Point
Clouds},
 year = {2021}
}

@inproceedings{chatscene,
title={Chat-3d v2: Bridging 3d scene and large language models with object identifiers},
  author={Huang, Haifeng and Wang, Zehan and Huang, Rongjie and Liu, Luping and Cheng, Xize and Zhao, Yang and Jin, Tao and Zhao, Zhou},
  booktitle=NeurIPS,
  year={2024}
}

@inproceedings{llava3d,
 title={Llava-3d: A simple yet effective pathway to empowering lmms with 3d-awareness},
  author={Zhu, Chenming and Wang, Tai and Zhang, Wenwei and Pang, Jiangmiao and Liu, Xihui},
  booktitle= ICCV,
  year={2024}
}

@inproceedings{leo,
 author = {Jiangyong Huang and
Silong Yong and
Xiaojian Ma and
Xiongkun Linghu and
Puhao Li and
Yan Wang and
Qing Li and
Song{-}Chun Zhu and
Baoxiong Jia and
Siyuan Huang},
 booktitle = ICML,
 title = {An Embodied Generalist Agent in 3D World},
 year = {2024}
}

@inproceedings{sqa3d,
 author = {Xiaojian Ma and
Silong Yong and
Zilong Zheng and
Qing Li and
Yitao Liang and
Song{-}Chun Zhu and
Siyuan Huang},
 booktitle = ICLR,
 title = {{SQA3D:} Situated Question Answering in 3D Scenes},
 year = {2023},
note={License: CC-BY-4.0}
}

@InProceedings{scenellm,
    author    = {Fu, Rao and Liu, Jingyu and Chen, Xilun and Nie, Yixin and Xiong, Wenhan},
    title     = {Scene-LLM: Extending Language Model for 3D Visual Reasoning},
    booktitle = {Proceedings of the Winter Conference on Applications of Computer Vision (WACV)},
    month     = {February},
    year      = {2025},
    pages     = {2195-2206}
}

@inproceedings{pointllm,
 author = {Runsen Xu and
Xiaolong Wang and
Tai Wang and
Yilun Chen and
Jiangmiao Pang and
Dahua Lin},
 booktitle = ECCV,
 title = {PointLLM: Empowering Large Language Models to Understand Point Clouds},
 year = {2024}
}

@inproceedings{scannet,
  title={Scannet: Richly-annotated 3d reconstructions of indoor scenes},
  author={Dai, Angela and Chang, Angel X and Savva, Manolis and Halber, Maciej and Funkhouser, Thomas and Nie{\ss}ner, Matthias},
  booktitle=CVPR,
  year={2017},
  note={License: ScanNet Terms of Use}
}

@inproceedings{pq3d,
  title={Unifying 3d vision-language understanding via promptable queries},
  author={Zhu, Ziyu and Zhang, Zhuofan and Ma, Xiaojian and Niu, Xuesong and Chen, Yixin and Jia, Baoxiong and Deng, Zhidong and Huang, Siyuan and Li, Qing},
  booktitle=ECCV,
  year={2024},
}

@inproceedings{radford2021learning,
  title={Learning transferable visual models from natural language supervision},
  author={Radford, Alec and Kim, Jong Wook and Hallacy, Chris and Ramesh, Aditya and Goh, Gabriel and Agarwal, Sandhini and Sastry, Girish and Askell, Amanda and Mishkin, Pamela and Clark, Jack and others},
  booktitle=ICML,
  pages={8748--8763},
  year={2021},
  organization={PMLR}
}

@inproceedings{kim2024openvla,
  title={OpenVLA: An Open-Source Vision-Language-Action Model},
  author={Kim, Moo Jin and Pertsch, Karl and Karamcheti, Siddharth and Xiao, Ted and Balakrishna, Ashwin and Nair, Suraj and Rafailov, Rafael and Foster, Ethan and Lam, Grace and Sanketi, Pannag and others},
  booktitle={Proceedings of the 8th Conference on Robot Learning},
  pages={2679--2713},
  year={2024},
  publisher={PMLR}
}

@article{chi2023diffusion,
  title={Diffusion policy: Visuomotor policy learning via action diffusion},
  author={Chi, Cheng and Xu, Zhenjia and Feng, Siyuan and Cousineau, Eric and Du, Yilun and Burchfiel, Benjamin and Tedrake, Russ and Song, Shuran},
  journal={The International Journal of Robotics Research},
  year={2023},
  publisher={SAGE Publications Sage UK: London, England}
}

@article{liu2024libero,
  title={Libero: Benchmarking knowledge transfer for lifelong robot learning},
  author={Liu, Bo and Zhu, Yifeng and Gao, Chongkai and Feng, Yihao and Liu, Qiang and Zhu, Yuke and Stone, Peter},
  journal={Advances in Neural Information Processing Systems},
  volume={36},
  year={2024}
}

@inproceedings{lipman2022flow,
  title={Flow matching for generative modeling},
  author={Lipman, Yaron and Chen, Ricky TQ and Ben-Hamu, Heli and Nickel, Maximilian and Le, Matt},
  booktitle=ICLR,
  year={2023}
}

@article{team2024octo,
  title={Octo: An open-source generalist robot policy},
  author={Team, Octo Model and Ghosh, Dibya and Walke, Homer and Pertsch, Karl and Black, Kevin and Mees, Oier and Dasari, Sudeep and Hejna, Joey and Kreiman, Tobias and Xu, Charles and others},
  journal={Robotics: Science and Systems (RSS)},
  year={2024}
}

@inproceedings{zhai2023sigmoid,
  title={Sigmoid loss for language image pre-training},
  author={Zhai, Xiaohua and Mustafa, Basil and Kolesnikov, Alexander and Beyer, Lucas},
  booktitle={Proceedings of the IEEE/CVF International Conference on Computer Vision},
  pages={11975--11986},
  year={2023}
}

@inproceedings{hou2025dita,
  title={Dita: Scaling diffusion transformer for generalist vision-language-action policy},
  author={Hou, Zhi and Zhang, Tianyi and Xiong, Yuwen and Duan, Haonan and Pu, Hengjun and Tong, Ronglei and Zhao, Chengyang and Zhu, Xizhou and Qiao, Yu and Dai, Jifeng and others},
  booktitle=ICCV,
  pages={7686--7697},
  year={2025}
}

@inproceedings{huang2025,
  title={3DRS: MLLMs Need 3D-Aware Representation Supervision for Scene Understanding},
  author={Xiaohu Huang and Jingjing Wu and Qunyi Xie and Kai Han},
  booktitle=NIPS,
  year={2025}
}

@inproceedings{kondratyuk2023videopoet,
  title={Videopoet: A large language model for zero-shot video generation},
  author={Kondratyuk, Dan and Yu, Lijun and Gu, Xiuye and Lezama, Jos{\'e} and Huang, Jonathan and Schindler, Grant and Hornung, Rachel and Birodkar, Vighnesh and Yan, Jimmy and Chiu, Ming-Chang and others},
  booktitle=ICML,
  year={2024}
}

@article{genie3,
  title         = {Genie 3: A New Frontier for World Models},
  author        = {Philip J. Ball and Jakob Bauer and Frank Belletti and Bethanie Brownfield and Ariel Ephrat and Shlomi Fruchter and Agrim Gupta and Kristian Holsheimer and Aleksander Holynski and Jiri Hron and Christos Kaplanis and Marjorie Limont and Matt McGill and Yanko Oliveira and Jack Parker-Holder and Frank Perbet and Guy Scully and Jeremy Shar and Stephen Spencer and Omer Tov and Ruben Villegas and Emma Wang and Jessica Yung and Cip Baetu and Jordi Berbel and David Bridson and Jake Bruce and Gavin Buttimore and Sarah Chakera and Bilva Chandra and Paul Collins and Alex Cullum and Bogdan Damoc and Vibha Dasagi and Maxime Gazeau and Charles Gbadamosi and Woohyun Han and Ed Hirst and Ashyana Kachra and Lucie Kerley and Kristian Kjems and Eva Knoepfel and Vika Koriakin and Jessica Lo and Cong Lu and Zeb Mehring and Alex Moufarek and Henna Nandwani and Valeria Oliveira and Fabio Pardo and Jane Park and Andrew Pierson and Ben Poole and Helen Ran and Tim Salimans and Manuel Sanchez and Igor Saprykin and Amy Shen and Sailesh Sidhwani and Duncan Smith and Joe Stanton and Hamish Tomlinson and Dimple Vijaykumar and Luyu Wang and Piers Wingfield and Nat Wong and Keyang Xu and Christopher Yew and Nick Young and Vadim Zubov and Douglas Eck and Dumitru Erhan and Koray Kavukcuoglu and Demis Hassabis and Zoubin Gharamani and Raia Hadsell and A{\"a}ron van den Oord and Inbar Mosseri and Adrian Bolton and Satinder Singh and Tim Rockt{\"a}schel},
  year          = {2025},
  url           = {https://deepmind.google/blog/genie-3-a-new-frontier-for-world-models/},
  note          = {Accessed: 28 June 2026}
}

@article{wan2025wan,
  title={Wan: Open and advanced large-scale video generative models},
  author={Wan, Team and Wang, Ang and Ai, Baole and Wen, Bin and Mao, Chaojie and Xie, Chen-Wei and Chen, Di and Yu, Feiwu and Zhao, Haiming and Yang, Jianxiao and others},
  journal={arXiv preprint arXiv:2503.20314},
  year={2025}
}

@inproceedings{ren2025gen3c,
  title={Gen3c: 3d-informed world-consistent video generation with precise camera control},
  author={Ren, Xuanchi and Shen, Tianchang and Huang, Jiahui and Ling, Huan and Lu, Yifan and Nimier-David, Merlin and M{\"u}ller, Thomas and Keller, Alexander and Fidler, Sanja and Gao, Jun},
  booktitle={Proceedings of the Computer Vision and Pattern Recognition Conference},
  pages={6121--6132},
  year={2025}
}

@inproceedings{kim2025videofrom3d,
  title={VideoFrom3D: 3D Scene Video Generation via Complementary Image and Video Diffusion Models},
  author={Kim, Geonung and Han, Janghyeok and Cho, Sunghyun},
  booktitle={Proceedings of the SIGGRAPH Asia 2025 Conference Papers},
  pages={1--11},
  year={2025}
}

@inproceedings{kang2024far,
  title={How far is video generation from world model: A physical law perspective},
  author={Kang, Bingyi and Yue, Yang and Lu, Rui and Lin, Zhijie and Zhao, Yang and Wang, Kaixin and Huang, Gao and Feng, Jiashi},
  booktitle=ICML,
  year={2025}
}

@inproceedings{valevski2024diffusion,
  title={Diffusion models are real-time game engines},
  author={Valevski, Dani and Leviathan, Yaniv and Arar, Moab and Fruchter, Shlomi},
  booktitle=ICLR,
  year={2025}
}

@inproceedings{xiao2025worldmem,
  title={Worldmem: Long-term consistent world simulation with memory},
  author={Xiao, Zeqi and Lan, Yushi and Zhou, Yifan and Ouyang, Wenqi and Yang, Shuai and Zeng, Yanhong and Pan, Xingang},
  booktitle=NIPS,
  year={2025}
}

@inproceedings{zheng2024towards,
  title={Towards learning a generalist model for embodied navigation},
  author={Zheng, Duo and Huang, Shijia and Zhao, Lin and Zhong, Yiwu and Wang, Liwei},
  booktitle=CVPR,
  pages={13624--13634},
  year={2024}
}

@article{liu2025embodied,
  title={Embodied navigation},
  author={Liu, Yunhao and Liu, Li and Zheng, Yawen and Liu, Yunhuai and Dang, Fan and Li, Ningbo and Ma, Ke},
  journal={Science China Information Sciences},
  volume={68},
  number={4},
  pages={1--39},
  year={2025},
  publisher={Springer}
}

@inproceedings{gao2024physically,
  title={Physically grounded vision-language models for robotic manipulation},
  author={Gao, Jensen and Sarkar, Bidipta and Xia, Fei and Xiao, Ted and Wu, Jiajun and Ichter, Brian and Majumdar, Anirudha and Sadigh, Dorsa},
  booktitle=ICRA,
  pages={12462--12469},
  year={2024},
  organization={IEEE}
}

@article{cheang2024gr,
  title={Gr-2: A generative video-language-action model with web-scale knowledge for robot manipulation},
  author={Cheang, Chi-Lam and Chen, Guangzeng and Jing, Ya and Kong, Tao and Li, Hang and Li, Yifeng and Liu, Yuxiao and Wu, Hongtao and Xu, Jiafeng and Yang, Yichu and others},
  journal={arXiv preprint arXiv:2410.06158},
  year={2024}
}

@inproceedings{wang2025ross3d,
  title={Ross3d: Reconstructive visual instruction tuning with 3d-awareness},
  author={Wang, Haochen and Zhao, Yucheng and Wang, Tiancai and Fan, Haoqiang and Zhang, Xiangyu and Zhang, Zhaoxiang},
  booktitle=ICCV,
  year={2025}
}

@inproceedings{wang2025vggt,
  title={Vggt: Visual geometry grounded transformer},
  author={Wang, Jianyuan and Chen, Minghao and Karaev, Nikita and Vedaldi, Andrea and Rupprecht, Christian and Novotny, David},
  booktitle=CVPR,
  pages={5294--5306},
  year={2025}
}

@article{chen2025think,
  title={Think with 3d: Geometric imagination grounded spatial reasoning from limited views},
  author={Chen, Zhangquan and Zhang, Manyuan and Yu, Xinlei and Luo, Xufang and Sun, Mingze and Pan, Zihao and Feng, Yan and Pei, Peng and Cai, Xunliang and Huang, Ruqi},
  journal={arXiv preprint arXiv:2510.18632},
  year={2025}
}

@inproceedings{huang2025vid2world,
  title={Vid2World: Crafting Video Diffusion Models to Interactive World Models},
  author={Huang, Siqiao and Wu, Jialong and Zhou, Qixing and Miao, Shangchen and Long, Mingsheng},
  booktitle=ICLR,
  year={2026}
}

@inproceedings{zhou2025stable,
  title={Stable virtual camera: Generative view synthesis with diffusion models},
  author={Zhou, Jensen and Gao, Hang and Voleti, Vikram and Vasishta, Aaryaman and Yao, Chun-Han and Boss, Mark and Torr, Philip and Rupprecht, Christian and Jampani, Varun},
  booktitle=ICCV,
  year={2025}
}

@inproceedings{li2025vmem,
  title={VMem: Consistent Interactive Video Scene Generation with Surfel-Indexed View Memory},
  author={Li, Runjia and Torr, Philip and Vedaldi, Andrea and Jakab, Tomas},
  booktitle=ICCV,
  year={2025}
}

@article{blattmann2023stable,
  title={Stable video diffusion: Scaling latent video diffusion models to large datasets},
  author={Blattmann, Andreas and Dockhorn, Tim and Kulal, Sumith and Mendelevitch, Daniel and Kilian, Maciej and Lorenz, Dominik and Levi, Yam and English, Zion and Voleti, Vikram and Letts, Adam and others},
  journal={arXiv preprint arXiv:2311.15127},
  year={2023}
}

@article{kim2025fine,
  title={Fine-tuning vision-language-action models: Optimizing speed and success},
  author={Kim, Moo Jin and Finn, Chelsea and Liang, Percy},
  journal={Robotics: Science and Systems (RSS)},
  year={2025}
}

@article{zhang2025dyva,
  title={Can World Models Benefit {VLMs} for World Dynamics?},
  author={Zhang, Kevin and Ge, Kuangzhi and Chi, Xiaowei and Zhang, Renrui and Shi, Shaojun and Dong, Zhen and Han, Sirui and Zhang, Shanghang},
  journal={arXiv preprint arXiv:2510.00855},
  year={2025}
}

@article{miao2026jepavla,
  title={{JEPA-VLA}: Video Predictive Embedding is Needed for {VLA} Models},
  author={Miao, Shangchen and Feng, Ningya and Wu, Jialong and Lin, Ye and He, Xu and Li, Dong and Long, Mingsheng},
  journal={arXiv preprint arXiv:2602.11832},
  year={2026}
}

@inproceedings{mu2025robotwin,
  title={{RoboTwin}: Dual-Arm Robot Benchmark with Generative Digital Twins},
  author={Mu, Yao and Chen, Tianxing and others},
  booktitle=CVPR,
  year={2025}
}

@article{assran2025v,
  title={V-jepa 2: Self-supervised video models enable understanding, prediction and planning},
  author={Assran, Mido and Bardes, Adrien and Fan, David and Garrido, Quentin and Howes, Russell and Muckley, Matthew and Rizvi, Ammar and Roberts, Claire and Sinha, Koustuv and Zholus, Artem and others},
  journal={arXiv preprint arXiv:2506.09985},
  year={2025}
}

@inproceedings{rombach2022high,
  title={High-resolution image synthesis with latent diffusion models},
  author={Rombach, Robin and Blattmann, Andreas and Lorenz, Dominik and Esser, Patrick and Ommer, Bj{\"o}rn},
  booktitle=CVPR,
  pages={10684--10695},
  year={2022}
}

@inproceedings{zhou2025llava,
  title={LLaVA-4D: Embedding SpatioTemporal Prompt into LMMs for 4D Scene Understanding},
  author={Zhou, Hanyu and Lee, Gim Hee},
  booktitle=ICLR,
  year={2026}
}

@inproceedings{zhang2025sphere,
  title={Sphere: Unveiling spatial blind spots in vision-language models through hierarchical evaluation},
  author={Zhang, Wenyu and Ng, Wei En and Ma, Lixin and Wang, Yuwen and Zhao, Junqi and Koenecke, Allison and Li, Boyang and Wanglu, Wanglu},
  booktitle=ACL,
  pages={11591--11609},
  year={2025}
}

@inproceedings{yang2025thinking,
  title={Thinking in space: How multimodal large language models see, remember, and recall spaces},
  author={Yang, Jihan and Yang, Shusheng and Gupta, Anjali W and Han, Rilyn and Fei-Fei, Li and Xie, Saining},
  booktitle=CVPR,
  pages={10632--10643},
  year={2025}
}

@article{chen2025cvp,
  title={CVP: Central-Peripheral Vision-Inspired Multimodal Model for Spatial Reasoning},
  author={Chen, Zeyuan and Zhang, Xiang and Xu, Haiyang and Xie, Jianwen and Tu, Zhuowen},
  journal={arXiv preprint arXiv:2512.08135},
  year={2025}
}

@inproceedings{yin2025spatial,
  title={Spatial mental modeling from limited views},
  author={Yin, Baiqiao and Wang, Qineng and Zhang, Pingyue and Zhang, Jianshu and Wang, Kangrui and Wang, Zihan and Zhang, Jieyu and Chandrasegaran, Keshigeyan and Liu, Han and Krishna, Ranjay and others},
  booktitle={Structural Priors for Vision Workshop at ICCV'25},
  year={2025}
}

@inproceedings{chen2024spatialvlm,
  title={Spatialvlm: Endowing vision-language models with spatial reasoning capabilities},
  author={Chen, Boyuan and Xu, Zhuo and Kirmani, Sean and Ichter, Brain and Sadigh, Dorsa and Guibas, Leonidas and Xia, Fei},
  booktitle=CVPR,
  pages={14455--14465},
  year={2024}
}

@inproceedings{fan2025vlm,
  title={VLM-3R: Vision-Language Models Augmented with Instruction-Aligned 3D Reconstruction},
  author={Fan, Zhiwen and Zhang, Jian and Li, Renjie and Zhang, Junge and Chen, Runjin and Hu, Hezhen and Wang, Kevin and Qu, Huaizhi and Wang, Dilin and Yan, Zhicheng and others},
  booktitle=CVPR,
  year={2026}
}

@article{liu2023visual,
  title={Visual instruction tuning},
  author={Liu, Haotian and Li, Chunyuan and Wu, Qingyang and Lee, Yong Jae},
  journal={Advances in neural information processing systems},
  volume={36},
  pages={34892--34916},
  year={2023}
}

@article{videoworldsimulators2024,
  title={Video generation models as world simulators},
  author={Tim Brooks and Bill Peebles and Connor Holmes and Will DePue and Yufei Guo and Li Jing and David Schnurr and Joe Taylor and Troy Luhman and Eric Luhman and Clarence Ng and Ricky Wang and Aditya Ramesh},
  year={2024},
  url={https://openai.com/research/video-generation-models-as-world-simulators},
  note={Accessed: 28 June 2026},
}

@inproceedings{yang2024cogvideox,
  title={CogVideoX: Text-to-Video Diffusion Models with An Expert Transformer},
  author={Yang, Zhuoyi and Teng, Jiayan and Zheng, Wendi and Ding, Ming and Huang, Shiyu and Xu, Jiazheng and Yang, Yuanming and Hong, Wenyi and Zhang, Xiaohan and Feng, Guanyu and others},
  booktitle=ICLR,
  year={2025}
}

@inproceedings{hong2022cogvideo,
  title={CogVideo: Large-scale Pretraining for Text-to-Video Generation via Transformers},
  author={Hong, Wenyi and Ding, Ming and Zheng, Wendi and Liu, Xinghan and Tang, Jie},
   booktitle=ICLR,
  year={2023}
}

@inproceedings{jia2025omnispatial,
  title={OmniSpatial: Towards Comprehensive Spatial Reasoning Benchmark for Vision Language Models},
  author={Jia, Mengdi and Qi, Zekun and Zhang, Shaochen and Zhang, Wenyao and Yu, Xinqiang and He, Jiawei and Wang, He and Yi, Li},
  booktitle=ICLR,
  year={2026}
}

@inproceedings{lin2025ost,
  title={Ost-bench: Evaluating the capabilities of mllms in online spatio-temporal scene understanding},
  author={Lin, JingLi and Zhu, Chenming and Xu, Runsen and Mao, Xiaohan and Liu, Xihui and Wang, Tai and Pang, Jiangmiao},
  booktitle=NeurIPS,
  year={2025}
}

@inproceedings{yang2025mmsi,
  title={MMSI-Bench: A Benchmark for Multi-Image Spatial Intelligence},
  author={Yang, Sihan and Xu, Runsen and Xie, Yiman and Yang, Sizhe and Li, Mo and Lin, Jingli and Zhu, Chenming and Chen, Xiaochen and Duan, Haodong and Yue, Xiangyu and others},
   booktitle=ICLR,
  year={2026}
}

@inproceedings{chen2025vl,
  title={Vl-jepa: Joint embedding predictive architecture for vision-language},
  author={Chen, Delong and Shukor, Mustafa and Moutakanni, Theo and Chung, Willy and Yu, Jade and Kasarla, Tejaswi and Bolourchi, Allen and LeCun, Yann and Fung, Pascale},
  booktitle=ICLR,
  year={2026}
}

@inproceedings{jiang2025vace,
  title={Vace: All-in-one video creation and editing},
  author={Jiang, Zeyinzi and Han, Zhen and Mao, Chaojie and Zhang, Jingfeng and Pan, Yulin and Liu, Yu},
  booktitle=ICCV,
  pages={17191--17202},
  year={2025}
}

@article{simeoni2025dinov3,
  title={Dinov3},
  author={Sim{\'e}oni, Oriane and Vo, Huy V and Seitzer, Maximilian and Baldassarre, Federico and Oquab, Maxime and Jose, Cijo and Khalidov, Vasil and Szafraniec, Marc and Yi, Seungeun and Ramamonjisoa, Micha{\"e}l and others},
  journal={arXiv preprint arXiv:2508.10104},
  year={2025}
}

@inproceedings{mei2026efficientencoderfreefourierbased3d,
      title={Efficient Encoder-Free Fourier-based 3D Large Multimodal Model}, 
      author={Guofeng Mei and Wei Lin and Luigi Riz and Yujiao Wu and Yiming Wang and Fabio Poiesi},
      year={2026},
      booktitle=CVPR,
}

@article{hurst2024gpt4o,
  title={Gpt-4o system card},
  author={Hurst, Aaron and Lerer, Adam and Goucher, Adam P and Perelman, Adam and Ramesh, Aditya and Clark, Aidan and Ostrow, AJ and Welihinda, Akila and Hayes, Alan and Radford, Alec and others},
  journal={arXiv preprint arXiv:2410.21276},
  year={2024}
}

@article{team2024gemini,
  title={Gemini 1.5: Unlocking multimodal understanding across millions of tokens of context},
  author={{Gemini Team}},
  journal={arXiv preprint arXiv:2403.05530},
  year={2024}
}

@article{chen2024internvl2,
  title={How far are we to gpt-4v? closing the gap to commercial multimodal models with open-source suites},
  author={Chen, Zhe and Wang, Weiyun and Tian, Hao and Ye, Shenglong and Gao, Zhangwei and Cui, Erfei and Tong, Wenwen and Hu, Kongzhi and Luo, Jiapeng and Ma, Zheng and others},
  journal={Science China Information Sciences},
  volume={67},
  number={12},
  pages={220101},
  year={2024},
  publisher={Springer}
}

@article{li2024llava-onevision,
  title={Llava-onevision: Easy visual task transfer},
  author={Li, Bo and Zhang, Yuanhan and Guo, Dong and Zhang, Renrui and Li, Feng and Zhang, Hao and Zhang, Kaichen and Zhang, Peiyuan and Li, Yanwei and Liu, Ziwei and others},
  journal={Transactions on Machine Learning Research},
  year={2025}
}

@misc{liu2024llavanext,
    title={LLaVA-NeXT: Improved reasoning, OCR, and world knowledge},
    url={https://llava-vl.github.io/blog/2024-01-30-llava-next/},
    author={Liu, Haotian and Li, Chunyuan and Li, Yuheng and Li, Bo and Zhang, Yuanhan and Shen, Sheng and Lee, Yong Jae},
    month={January},
    year={2024},
    note={Accessed: 28 June 2026}
}

@inproceedings{zheng2025learning,
  title={Learning from Videos for 3D World: Enhancing MLLMs with 3D Vision Geometry Priors},
  author={Zheng, Duo and Huang, Shijia and Li, Yanyang and Wang, Liwei},
  booktitle=NIPS,
  year={2025}
}

@article{Qwen2.5-VL,
  title={Qwen2.5-VL Technical Report},
  author={{Qwen Team}},
  journal={arXiv preprint arXiv:2502.13923},
  year={2025}
}

@inproceedings{zhang2025flatland,
  title={From flatland to space: Teaching vision-language models to perceive and reason in 3d},
  author={Zhang, Jiahui and Chen, Yurui and Zhou, Yanpeng and Xu, Yueming and Huang, Ze and Mei, Jilin and Chen, Junhui and Yuan, Yu-Jie and Cai, Xinyue and Huang, Guowei and others},
  booktitle=ICCV,
  year={2025}
}

@inproceedings{liu2025nvila,
  title={Nvila: Efficient frontier visual language models},
  author={Liu, Zhijian and Zhu, Ligeng and Shi, Baifeng and Zhang, Zhuoyang and Lou, Yuming and Yang, Shang and Xi, Haocheng and Cao, Shiyi and Gu, Yuxian and Li, Dacheng and others},
  booktitle=CVPR,
  pages={4122--4134},
  year={2025}
}

@article{zhang2024long,
  title={Long context transfer from language to vision},
  author={Zhang, Peiyuan and Zhang, Kaichen and Li, Bo and Zeng, Guangtao and Yang, Jingkang and Zhang, Yuanhan and Wang, Ziyue and Tan, Haoran and Li, Chunyuan and Liu, Ziwei},
  journal={arXiv preprint arXiv:2406.16852},
  year={2024}
}

@inproceedings{chen2024longvila,
  title={Longvila: Scaling long-context visual language models for long videos},
  author={Chen, Yukang and Xue, Fuzhao and Li, Dacheng and Hu, Qinghao and Zhu, Ligeng and Li, Xiuyu and Fang, Yunhao and Tang, Haotian and Yang, Shang and Liu, Zhijian and others},
  booktitle=ICLR,
  year={2025}
}

@inproceedings{feng2025video,
  title={Video-r1: Reinforcing video reasoning in mllms},
  author={Feng, Kaituo and Gong, Kaixiong and Li, Bohao and Guo, Zonghao and Wang, Yibing and Peng, Tianshuo and Wu, Junfei and Zhang, Xiaoying and Wang, Benyou and Yue, Xiangyu},
  booktitle=NIPS,
  year={2025}
}

@article{liao2025improved,
  title={Improved visual-spatial reasoning via r1-zero-like training},
  author={Liao, Zhenyi and Xie, Qingsong and Zhang, Yanhao and Kong, Zijian and Lu, Haonan and Yang, Zhenyu and Deng, Zhijie},
  journal={arXiv preprint arXiv:2504.00883},
  year={2025}
}

@article{ouyang2025spacer,
  title={Spacer: Reinforcing mllms in video spatial reasoning},
  author={Ouyang, Kun and Liu, Yuanxin and Wu, Haoning and Liu, Yi and Zhou, Hao and Zhou, Jie and Meng, Fandong and Sun, Xu},
  journal={arXiv preprint arXiv:2504.01805},
  year={2025}
}

@article{xu2024unified,
  title={A unified framework for 3d scene understanding},
  author={Xu, Wei and Shi, Chunsheng and Tu, Sifan and Zhou, Xin and Liang, Dingkang and Bai, Xiang},
  journal={Advances in Neural Information Processing Systems},
  volume={37},
  pages={59468--59490},
  year={2024}
}

@article{UniVLA,
  title={Univla: Learning to act anywhere with task-centric latent actions},
  author={Bu, Qingwen and Yang, Yanting and Cai, Jisong and Gao, Shenyuan and Ren, Guanghui and Yao, Maoqing and Luo, Ping and Li, Hongyang},
  journal={Robotics: Science and Systems (RSS)},
  year={2025}
}

@inproceedings{CoT-VLA-2025,
  title={Cot-vla: Visual chain-of-thought reasoning for vision-language-action models},
  author={Qingqing Zhao and Yao Lu and Moo Jin Kim and Zipeng Fu and Zhuoyang Zhang and Yecheng Wu and Zhaoshuo Li and Qianli Ma and Song Han and Chelsea Finn and Ankur Handa and Ming-Yu Liu and Donglai Xiang and Gordon Wetzstein and Tsung-Yi Lin},
  booktitle=CVPR,
  pages={1702--1713},
  year={2025}
}

@inproceedings{hu2025omniview,
  title={Omni-View: Unlocking How Generation Facilitates Understanding in Unified 3D Model based on Multiview images}, 
  author={JiaKui Hu and Shanshan Zhao and Qing-Guo Chen and Xuerui Qiu and Jialun Liu and Zhao Xu and Weihua Luo and Kaifu Zhang and Yanye Lu},
  year={2026},
  booktitle=ICLR,
}

@misc{cao2025seeingimaginationlearningscene,
      title={Seeing through Imagination: Learning Scene Geometry via Implicit Spatial World Modeling}, 
      author={Meng Cao and Haokun Lin and Haoyuan Li and Haoran Tang and Rongtao Xu and Dong An and Xue Liu and Ian Reid and Xiaodan Liang},
      year={2025},
      eprint={2512.01821},
      archivePrefix={arXiv},
      primaryClass={cs.CV},
      url={https://arxiv.org/abs/2512.01821},
      note={Accessed: 28 June 2026},
}

@article{xu2026nextforcing,
  title={Next Forcing: Causal World Modeling with Multi-Chunk Prediction},
  author={Xu, Gangwei and Zhang, Qihang and Zhou, Jiaming and Zhu, Xing and Shen, Yujun and Yang, Xin and Xu, Yinghao},
  journal={arXiv preprint arXiv:2606.11187},
  year={2026},
}

@article{deng2025best3dscenerepresentation,
  title={What Is The Best 3D Scene Representation for Robotics? From Geometric to Foundation Models},
  author={Deng, Tianchen and Pan, Yue and Yuan, Shenghai and Li, Dong and Wang, Chen and Li, Mingrui and Chen, Long and Xie, Lihua and Wang, Danwei and Wang, Jingchuan and Civera, Javier and Wang, Hesheng and Chen, Weidong},
  journal={arXiv preprint arXiv:2512.03422},
  year={2025},
}

@inproceedings{Deng_2026_CVPR,
  title={GaussianDWM: 3D Gaussian Driving World Model for Unified Scene Understanding and Multi-Modal Generation},
  author={Deng, Tianchen and Chen, Xuefeng and Chen, Yi and Chen, Qu and Xu, Yuyao and Yang, Lijin and Xu, Le and Zhang, Yu and Zhang, Bo and Huang, Wuxiong and Wang, Hesheng},
  booktitle=CVPR,
  year={2026},
}

@inproceedings{Zhang_2026_CVPR,
  title={OmniDrive-R1: Reinforcement-driven Interleaved Multi-modal Chain-of-Thought for Trustworthy Vision-Language Autonomous Driving},
  author={Zhang, Zhenguo and Zheng, Haohan and Wang, Yishen and Xu, Le and Deng, Tianchen and Chen, Xuefeng and Chen, Qu and Zhang, Bo and Huang, Wuxiong},
  booktitle=CVPR,
  year={2026},
}

@article{fang2026towards,
  title={Towards Generalizable Robotic Manipulation in Dynamic Environments},
  author={Fang, Heng and Li, Shangru and Wang, Shuhan and Xi, Xuanyang and Liang, Dingkang and Bai, Xiang},
  journal={arXiv preprint arXiv:2603.15620},
  year={2026},
}

@article{guan2026videostreamingthinking,
  title={Video Streaming Thinking: VideoLLMs Can Watch and Think Simultaneously},
  author={Guan, Yiran and Yin, Liang and Liang, Dingkang and Ju, Jianzhong and Luo, Zhenbo and Luan, Jian and Liu, Yuliang and Bai, Xiang},
  journal={arXiv preprint arXiv:2603.12262},
  year={2026},
}

@article{fu2025minddrive,
  title={MindDrive: A Vision-Language-Action Model for Autonomous Driving via Online Reinforcement Learning},
  author={Fu, Haoyu and Zhang, Diankun and Zhao, Zongchuang and Cui, Jianfeng and Xie, Hongwei and Wang, Bing and Chen, Guang and Liang, Dingkang and Bai, Xiang},
  journal={arXiv preprint arXiv:2512.13636},
  year={2025},
}
